\RequirePackage{fix-cm}
\documentclass[twocolumn]{svjour3}          
\smartqed  
\usepackage{tabularx}
\usepackage{rotating}
\usepackage{color}
\usepackage{multirow, booktabs}
\usepackage{graphicx}
\usepackage{boldline}
\usepackage{amsmath,amssymb,amsfonts, bm, mathtools}
\usepackage{float}
\usepackage{amsfonts}
\usepackage{natbib} 
\usepackage{threeparttable}
\usepackage{soul}

\usepackage{hyperref}

\usepackage[pdftex,dvipsnames]{xcolor}  
\usepackage[colorinlistoftodos,prependcaption, textsize=small, textwidth=30]{todonotes}

\usepackage{pgfplotstable}
\pgfplotsset{compat=1.12}
\usetikzlibrary{patterns}
\usepgfplotslibrary{groupplots}

\newcolumntype{P}[1]{>{\centering\arraybackslash}p{#1}}

\definecolor{Mulberry}{rgb}{0.77,0.29,0.55}
\definecolor{CadmiumOrange}{rgb}{0.93,0.53, 0.18}
\definecolor{ForestGreen}{rgb}{0.13, 0.55, 0.13}
\definecolor{WildStrawberry}{rgb}{0.5, 0.7, 0.2}

\newcommand{\miao}[1]{{\color{ForestGreen}{\bf{[Miao:]}} #1}}
\newcommand{\rui}[1]{{\color{WildStrawberry}{\bf{[Rui:]}} #1}}

\newtheorem{defn}{\hspace{-1mm} Definition}
\newcommand{\eat}[1]{}
\newcommand{\smallvspace}{\vspace{1.5mm}}
\newcommand{\smallhspace}{\hspace{0.6mm}}
\setcitestyle{square, aysep={,}}

\newcommand{\squishlist}{
   \begin{list}{$\bullet$}
    { \setlength{\itemsep}{0pt}      \setlength{\parsep}{3pt}
      \setlength{\topsep}{3pt}       \setlength{\partopsep}{0pt}
      \setlength{\leftmargin}{1.5em} \setlength{\labelwidth}{1em}
      \setlength{\labelsep}{0.5em} } }

\newcommand{\squishlisttwo}{
   \begin{list}{$\bullet$}
    { \setlength{\itemsep}{0pt}    \setlength{\parsep}{0pt}
      \setlength{\topsep}{0pt}     \setlength{\partopsep}{0pt}
      \setlength{\leftmargin}{2em} \setlength{\labelwidth}{1.5em}
      \setlength{\labelsep}{0.5em} } }

\newcommand{\squishend}{
    \end{list}  }

\begin{document}\sloppy

\title{A Benchmark and Comprehensive Survey on Knowledge Graph Entity Alignment via Representation Learning
}



\author{Rui Zhang \and Bayu Distiawan Trisedya \and Miao Li \and Yong Jiang \and Jianzhong Qi
}


\institute{Rui Zhang \at URL: \url{https://ruizhang.info/} \\
              \email{rayteam@yeah.net}              
          \and
          Bayu Distiawan Trisedya, Jianzhong Qi, Miao Li \at The University of Melbourne \\
              \email{\{bayu.trisedya, jianzhong.qi\}@unimelb.edu.au, miao4@student.unimelb.edu.au}
          \and
          Yong Jiang \at Tsinghua University \\
              \email{jiangy@mail.sz.tsinghua.edu.cn}
}

\date{Received: date / Accepted: date}

\maketitle

\begin{abstract}
In the last few years, the interest in knowledge bases has grown exponentially in both the research community and the industry due to their essential role in AI applications. Entity alignment is an important task for enriching knowledge bases. This paper provides a comprehensive tutorial-type survey on representative entity alignment techniques that use the new approach of representation learning. We present a framework for capturing the key characteristics of these techniques, propose a benchmark addressing the limitation of existing benchmark datasets, and conduct extensive experiments using our benchmark. The framework gives a clear picture of how various techniques work. The experiments yield important results about the empirical performance of the techniques and how various factors affect the performance. One important observation not stressed by previous work is that techniques making good use of attribute triples and relation predicates as features stand out as winners. We are also the first to investigate the question of how to perform entity alignments on large scale knowledge graphs such as the full Wikidata and Freebase (in Experiment 5).

\end{abstract}

\section{Introduction}

       \textit{Knowledge bases} are a technology used to store complex structured and unstructured information, typically facts or knowledge. A \textit{knowledge graph} (KG), which is a knowledge base modeled by a graph structure or topology, is the most popular form of knowledge bases and has almost become a synonym of knowledge base today. There have been continuous research and development on KGs for several decades due to their significance in systems that involve reasoning based on knowledge and facts. Example KGs include open-source ones such as DBpedia~\citep{auer2007dbpedia}, Freebase~\citep{bollacker2008freebase}, YAGO~\citep{hoffart2013yago2}, as well as proprietary ones such as those developed by Google~\citep{dong2014knowledge} and Microsoft~\citep{farber2019microsoft}. In the last few years, there has been an explosive growth of interest in KGs in both the research community and the industry due to their essential role in AI applications such as natural language processing (including dialogue systems/chatbots, question answering, sentence generation, etc.)~\citep{wu2017image,
       xu2019end,YangZE20,YangZEL21,gtr,GCP}, search engines~\citep{kathuria2016journey}, recommendation systems~\citep{zhang2016rec}, and information extraction~\citep{du2015using,TrisedyaWQZ19}.

       One of the most important tasks for KGs is \textit{entity alignment} (EA), which aims to identify entities from different KGs that represent the same real-world entity. EA enables enrichment of a KG from another complementary one, hence improving the quality and coverage of the KG, which is critical for downstream applications. Different KGs may be created via different sources and methods, so even entities representing the same real-world entity may be denoted differently in different KGs, and it is challenging to identify all such \textit{aligned entities} accurately.  { {Figure~\ref{fig:ea_example} %
       \begin{figure}[htp]
       \centering
       \includegraphics[width=0.9\columnwidth]{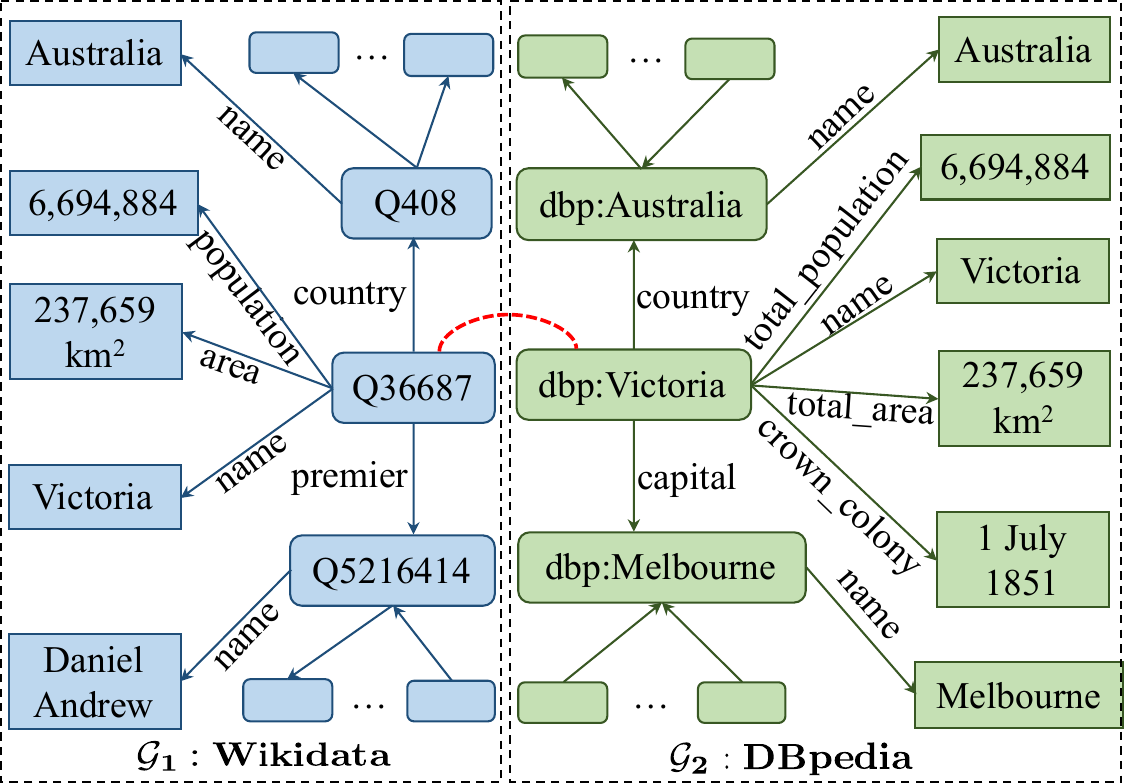}
       \caption{An example of EA.}
       \label{fig:ea_example}
       \vspace{-3mm}
       \end{figure}
       shows a toy example of EA on two KGs $\mathcal{G}_1$ and $\mathcal{G}_2$ (each in a dashed-line rectangle), which are tiny subsets from two real publicly available KGs, Wikidata and DBpedia, respectively. The rounded rectangles represent entities and the rectangles represent attribute values. An arrow between rounded rectangles indicates a relation predicate, which forms a \textit{relation triple}, e.g., (\texttt{dbp:Victoria, country, dbp:Australia}). An arrow between a rounded rectangle and a rectangle indicates an attribute predicate, which forms an \textit{attribute triple}, e.g., (\texttt{dbp:Victoria, total\_area, "237,659 km$^2$"}). We can see that the same real-world entity may have different surface forms in the two KGs such as \texttt{Q36687} v.s. \texttt{dbp:Victoria}. The two KGs have complementary information about this entity: $\mathcal{G}_1$ has information about its premier and $\mathcal{G}_2$ has information about its capital. The information about this entity can be enriched if we can determine that \texttt{Q36687} in $\mathcal{G}_1$ refers to the same real-world entity as \texttt{dbp:Victoria} in $\mathcal{G}_2$, i.e., \texttt{Q36687} and \texttt{dbp:Victoria} are \textit{aligned entities}. EA between $\mathcal{G}_1$ and $\mathcal{G}_2$ is to find all the pairs of aligned entities from the two KGs. In this example, there are two pairs of aligned entities $\langle \texttt{Q36687}, \texttt{dbp:Victoria}\rangle$ and $\langle \texttt{Q408}, \texttt{dbp:Australia}\rangle$.}}
       
       Traditional EA techniques use data mining or database approaches, typically heuristics, to identify similar entities. The accuracy of such approaches are limited, and heuristics are difficult to generalize. In the past several years, a very large number of studies on EA take the new approach of deep learning to learn effective vector representation (i.e., embeddings) of the KG and then performing EA based on the learned representation, which achieve much better accuracy. They also have better generalizability as they rarely rely on ad hoc heuristics. In the rest of this paper, by saying \textit{embedding-based EA techniques} or simply \textit{EA techniques}, we refer to those taking this new representation learning approach rather than traditional approaches unless explicitly specified otherwise. There are a few recent experimental studies aiming at benchmarking EA techniques \citep{sun2020benchmarking, zhao2020experimental, zhang2020an}. They have  high-level discussions on frameworks for embedding-based techniques and summarize a good range of EA papers, but their focus is on experimental comparison, but without self-contained explanation on each technique.  { {Moreover, the frameworks discussed in those papers miss important mechanisms such as the use of semantic information of KGs (e.g., strings of relation predicates, attribute predicates and attribute values), making those frameworks inapplicable to many EA techniques, especially the latest ones. In comparison to the aforementioned studies, this paper fills the void and make the following contributions:
       \squishlist
       \item We provide a comprehensive \textit{tutorial-type} survey to help readers understand how each technique works with little need to refer to individual full papers.
       \item We provide a comprehensive framework that accommodates almost all the embedding-based EA techniques, capturing their key components, strategies and characteristics. We also comparatively analyze different techniques in reference to the framework.
       \item We identify significant limitations of existing benchmark datasets such as bijection, lack of name variety, and small scale (detailed in Section~\ref{sec:dataset-limitation}). To address these limitations, we devise a benchmark\footnote{Our benchmark and all the code for our experiments are available at~\url{https://github.com/ruizhang-ai/EA\_for\_KG}}
       that complement the existing collection of benchmark datasets. Further, we conduct an extensive experimental study comparing the performance of the state-of-the-art techniques on our datasets. 
       \squishend
       }}
       
       The rest of the paper is organized as follows. Section~\ref{sec:pre} provides preliminaries, including problem formulation and a summary on traditional EA techniques. We present our framework for EA techniques in Section~\ref{sec:framework}. Section~\ref{sec:embedding} covers KG structure embedding models, mainly \textit{translation-based} and \textit{graph-neural-network-based} embedding, which are the foundation of embedding-based EA techniques.  {{Sections~\ref{sec:translationmodels} and \ref{sec:gnnmodels} survey the most representative EA techniques based on the two major KG structure embeddings, respectively.}} Section~\ref{sec:experiments} discusses the limitations of existing datasets, presents our proposed new datasets, and reports an extensive experimental study using our datasets. Section~\ref{sec:conclusion} concludes the paper and discusses future directions.

\section{Preliminaries}\label{sec:pre}
     \textbf{Notation and Terminology.} Many different notation and terminology conventions have been used in different papers in the literature. In this paper, we make a great effort at a standard notation and terminology convention that provides clarity and is consistent with as many existing papers as possible. The terminology will be seen throughout the paper as various terms are introduced, and the frequently used symbols in our notation convention are summarized in Table~\ref{tab:symbols}.
      \begin{table}[htp]
      \vspace{-2mm}
        \caption{Frequently Used Symbols}
        \setlength{\tabcolsep}{2.5pt}
        \centering
        \begin{tabular}{lp{60mm}}
        \toprule[2pt]
        \midrule
        \textbf{Symbols} & \textbf{Descriptions} \\
        \midrule
        $(\mathcal{E}, \mathcal{R}, \mathcal{A}, \mathcal{V}, \mathcal{T})$ & A knowledge graph\\
        \midrule
        $\mathcal{E}$ & A set of entities\\
        \midrule
        $\mathcal{R}$ & A set of relation predicates\\
        \midrule
        $\mathcal{A}$ & A set of attribute predicates\\
        \midrule
        $\mathcal{V}$ & A set of attribute values which may be numeric or literal\\
        \midrule
        $\mathcal{T}$ & A set of triples, which may consist of relation triples $\mathcal{T}_{r}$ and attribute triples $\mathcal{T}_{a}$\\
        \midrule
        $(h, r, t)$ & A relation triple, which consists of a head entity $h$, a tail entity $t$, the relation predicate $r$ between the entities\\
        \midrule
        $\boldsymbol{h}, \boldsymbol{r}, \boldsymbol{t}$ & Embeddings of the head entity, relation predicate and tail entity, respectively\\
        \midrule
        $(e, a, v)$ & An attribute triple, which consists of an entity $e$, an attribute predicate $a$ and the attribute value $v$\\
        \midrule
        $\boldsymbol{e}, \boldsymbol{a}, \boldsymbol{v}$ & Embeddings of an entity, attribute predicate and attribute value, respectively\\
        \midrule
        $f_\text{triple}$ & Triple score function\\
        \midrule
        $f_\text{align}$ & Alignment score function\\
        \midrule
        $\sigma$ & An non-linear activation function\\
        \midrule
        $\mathcal{S}$ & A set of pre-aligned entities from $\mathcal{G}_1$ and $\mathcal{G}_2$\\
        \midrule
        $\mathcal{S^\prime}$ & A set of corrupted entity alignments from $\mathcal{G}_1$ and $\mathcal{G}_2$\\
        \midrule
        $\mathcal{T}_r^\prime$ & A set of corrupted relation triples\\
        \midrule
        $\|$ & Concatenation of two vectors\\
        \midrule
        \bottomrule[2pt]
        \end{tabular}
        \label{tab:symbols}
        \vspace{-2mm}
    \end{table}
     We use bold lowercase (e.g., $\boldsymbol{e}$), bold uppercase (e.g., $\boldsymbol{M}$) and math calligraphy (e.g., $\mathcal{E}$) to denote vectors, matrices, and sets, respectively.
     
     In the literature, a method proposed by a paper may have been referred to by different terms such as model, approach, technique, algorithm, method, etc.; we primarily use the term \textit{technique} in this paper, while other terms might be used when the semantics are clear.

    \subsection{Problem Formulation}

     {We { first introduce some notation. A KG denoted as $\mathcal{G}=(\mathcal{E}, \mathcal{R}, \mathcal{A}, \mathcal{V}, \mathcal{T})$, consisting of a set of entities $\mathcal{E}$, a set of relation predicates $\mathcal{R}$, a set of attribute predicates $\mathcal{A}$, and a set of attribute values $\mathcal{V}$, represents knowledge in the form of a set of triples $\mathcal{T}$. There are two types of triples, \textit{relation triples} (denoted by $\mathcal{T}_{r}$) in the form of $(h,r,t)$ and \textit{attribute triples} (denoted by $\mathcal{T}_{a}$) in the form of $(e,a,v)$; $\mathcal{T} = \mathcal{T}_{r} \cup \mathcal{T}_{a}$. A relation triple $(h,r,t)$ indicates a relation predicate $r$ between two entities, a head entity $h$ and a tail entity $t$, where $h,t \in \mathcal{E}$ and $r \in \mathcal{R}$. Take a triple in Fig~\ref{fig:ea_example} as an example: (\texttt{dbp:Victoria, country, dbp:Australia}). Here, $\texttt{dbp:Victoria}$. and $\texttt{dbp:Australia}$ are the head entity and tail entity, respectively, and $\texttt{county}$ is the relation predicate. An attribute triple $(e,a,v)$ indicates that an entity $e \in \mathcal{E}$ has the attribute value of $ v \in \mathcal{V}$ for the attribute (predicate) $a \in \mathcal{A}$. For example, in (\texttt{dbp:Victoria, total\_area, "237,659 km$^2$"}),  $\texttt{total\_area}$ is the attribute predicate and $\texttt{"237,659 km$^2$"}$ is the attribute value.
    
    The problem of EA is formulated as follows.
    \vspace{-0mm}
    \begin{defn}{\textbf{Entity Alignment (EA)}}\\ Given two KGs $\mathcal{G}_1=(\mathcal{E}_1, \mathcal{R}_1, \mathcal{A}_1, \mathcal{V}_1, \mathcal{T}_1)$ and $\mathcal{G}_2=(\mathcal{E}_2, \mathcal{R}_2, \mathcal{A}_2, \mathcal{V}_2, \mathcal{T}_2)$, EA aims to identify every pair of entities $(e_1, e_2), e_1 \in \mathcal{E}_1, e_2 \in \mathcal{E}_2$, where $e_1$ and $e_2$ represent the same real-world entity (i.e., $e_1$ and $e_2$ are aligned entities). $\Box$
    \end{defn}
    \vspace{-2mm}
    }}

\subsection{Related Problems and Traditional Techniques}
\textbf{Related Problems}. There have been research on various problems similar to EA on KGs.

\underline{Both sources structured}. \textit{Entity matching}~\citep{verykios2000automating,Gcrewe2020}, \textit{object identification}~\citep{tejada2001learning}, and \textit{record linkage}~\citep{fellegi1969theory} aim to align entities from two different relational databases, where both data sources are well structured. Solutions for these problems mostly find database records that are similar in terms of contents.

\ul{One source semi-structured and the other unstructured}. \textit{Entity resolution}~\citep{bhattacharya2006entity} and \textit{entity linking}~\citep{kulkarni2009collective} aim to match entity mentions from natural sentences, which are unstructured, to the corresponding entities in a KG, which is semi-structured. A KG is semi-structured because it consists of a graph (structured), and attributes and predicates, which are in the form of natural language or other un-predefined types (unstructured). Solutions for these problems mostly find database records that have similar contents to named entities recognized from natural sentences.

In comparison, EA on KGs aligns entities from two different KGs, both of which are semi-structured.


\smallvspace
\noindent\textbf{Traditional Techniques for EA}.
Among traditional techniques for EA on KGs, some have focused on improving the \textit{effectiveness} of the matching of entities via different entity similarity measures. For example, RDF-AI~\citep{scharffe2009rdf} uses fuzzy string matching based on sequence alignment, 
 word relation,
and taxonomic similarity. SILK~\citep{volz2009discovering} provides the Link Specification Language, which allows users to specify the similarity measures for comparing certain attributes. LD-Mapper~\citep{raimond2008automatic} combines string similarity and entity nearest neighbors. 
PARIS \citep{suchanek2011paris} includes schema matching (e.g., classes and sub-classes of entities) to compute the entity similarity. Some other traditional EA techniques focus on the \textit{efficiency} of entity matching, e.g., LIMES~\citep{ngomo2011limes} uses clustering to reduce the amount of similarity computation. 

Traditional EA techniques, as exemplified above, usually use data mining or database approaches, typically heuristics, to identify similar entities. It is difficult for them to achieve high accuracy and to generalize.
\section{Generic Framework of Embedding-based EA}\label{sec:framework}
\begin{figure}[htp]
\vspace{-0mm}
    \centering
    \includegraphics[width=0.5\textwidth,trim=175 288 165 290, clip]{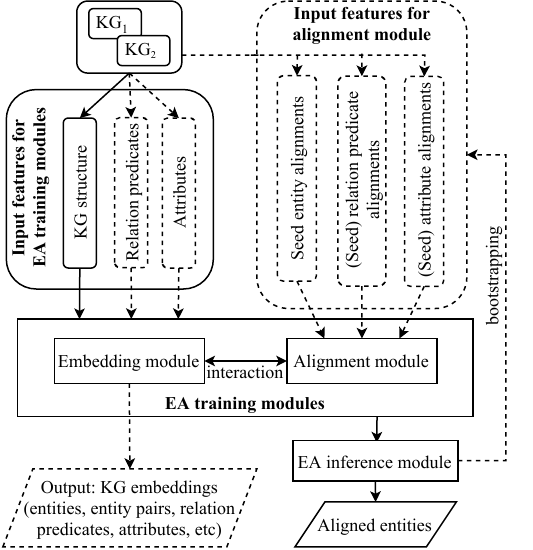}
    \caption{Framework of embedding-based EA techniques. Dashed lines indicate optional parts.}
    \vspace{-0mm}
    \label{fig:framework}
\end{figure}

We provide a generic framework for embedding-based EA techniques to capture key components and strategies in Figure~\ref{fig:framework}. The components drawn in dashed lines are optional. The approach of embedding-based EA typically consists of three components, an \emph{embedding module}, an \emph{alignment module}, and an \emph{inference module}. 
The embedding module and the alignment module may be trained alternatively or jointly, and these two together compose the \emph{training modules} for EA.

\smallvspace
\noindent \textbf{Embedding module}. The embedding module aims to learn (typically low-dimensional) vector representations, i.e., \textit{embeddings} of entities. There are four types of raw information that may be taken as input features to the embedding module: \textit{KG structure} (in the form of relation triples in the raw KG data), \textit{relation predicates}, \textit{attribute predicates} and \textit{attribute values} (attribute predicates and attribute values are grouped into one component ``Attributes" in Figure~\ref{fig:framework} and the 5$^{th}$ column in Table~\ref{tab:model_summary}).
The embedding module may produce as output the embeddings of entities, entity pairs, relation predicates, attributes, etc.; we refer to the process of ``encoding" the input features into the targeted embeddings as \textit{KG embedding}. 
Among all the possible input features, the KG structure is the most critical one. The machine learning model for embedding KG structure, which we simply term as the \textit{KG structure embedding model}, serves as the skeleton of an EA technique, and other types of information may be optionally added to the KG structure embedding model to create a more sophisticated KG embedding. Note that the other types of information (i.e., relation predicates, attribute predicates, and attribute values) are usually in the form of strings and hence contain rich semantic information, which may greatly benefit EA as we will see.

The KG structure embedding model mostly follows one of two paradigms, translation-based and GNN-based. Translation-based models mainly utilize relation triples while GNN-based models mainly utilize the neighborhood of entities. How relation triples are utilized in the translation-based models and how the neighborhood of entities is utilized in the GNN-based models are detailed in Sections~\ref{sec:transembedding} and \ref{sec:gnn-embedding}, respectively. 

Relation predicates and attribute predicates may be encoded as categorical values or strings. Attribute values are usually encoded as strings.


\smallvspace
\noindent 
\textbf{Alignment module}. The embedding module computes the embeddings of each KG separately, which makes the embeddings of $\mathcal{G}_1$ and $\mathcal{G}_2$ fall into different vector spaces. The alignment module aims to unify the embeddings of the two KGs into the same vector space so that aligned entities can be identified, which is a major challenge for EA. EA techniques usually make use of a set of manually aligned entities, relation predicates, or attributes, called \textit{seed alignments}, as input features to train the alignment module. The most common approach is using a set of seed \textit{entity} alignments $\mathcal{S} = \left\{(e_1,e_2)|e_1 \in \mathcal{E}_1, e_2 \in \mathcal{E}_2, e_1 \equiv e_2\right\}$. These seeds consist of pairs of entities $(e_1, e_2)$, where $e_1$ is an entity in $\mathcal{E}_1$ and $e_2$ is an entity in $\mathcal{E}_2$. The seeds are used to compute a loss function for the embedding module to learn a unified vector space. A typical example of how the loss function may be defined is as follows:
\begin{equation}
    \label{margin-based-kga-loss}
    \resizebox{.92\hsize}{!}{$\mathcal{L}=\sum\limits_{\left(e_1, e_2\right) \in \mathcal{S}} \sum\limits_{(e_1^\prime, e_2^\prime)\in \mathcal{S}^\prime}\max\Big(0, \big[\gamma + f_\text{align}(\boldsymbol{e}_1,\boldsymbol{e}_2)-f_\text{align}(\boldsymbol{e}_1^\prime,\boldsymbol{e}_2^\prime)\big]\Big)$}
\end{equation}
where $\gamma>0$ is a margin hyper-parameter. The above loss function is designed to minimize the distances between pairs of entities in the seed entity alignments $\mathcal{S}$, and maximize the distances between the pairs of entities $(e_1^\prime, e_2^\prime)$ in corrupted seed alignments $\mathcal{S}^\prime$, which are negative samples obtained by replacing an entity in the seed alignments with a random entity. Here, the distance between a pair of entities is computed by $f_\text{align}$, which we call the \emph{alignment score function}. It indicates how (dis)similar two entities are; the more the two entities are aligned (i.e., similar to each other), the smaller $f_\text{align}$ is.  {The most commonly used alignment score functions are cosine similarity, $L_1$ norm (i.e., Manhattan distance), and $L_2$ norm (i.e., Euclidean distance).
 { We do not observe large difference in performance in our experiments when swapping these metrics (also note that $L_2$ norm and cosine similarity are equivalent).}}
Some techniques customize the alignment score function to serve more sophisticated optimization goals such as~\cite{GCN-Align2018}.
The function $\max(0,)$ ensures that any negative margin loss values are not added to the total loss.
Some techniques may exploit other types of seed alignments, including \textit{seed relation predicate alignments}, \textit{seed attribute predicate alignments} and \textit{seed attribute value alignments} (seed attribute predicate alignments and seed attribute value alignments are grouped into one component ``Seed attribute alignment" in Figure~\ref{fig:framework}). Relation predicate and attribute predicate alignment are needed because the same predicate may be stored in different surface forms, e.g., one KG has the attribute predicate \texttt{birth\_date} while the other KG has the attribute predicate \texttt{date\_of\_birth}. The need for attribute value alignment is similar.

Note that like the embedding module, the alignment module may also use the four types of raw information (KG structure, relation predicates, attribute predicates, and attribute values) besides seed alignments as features. Some EA techniques may use an unsupervised method to train the alignment module; e.g., AttrE~\citep{ACE2019} exploits attribute triples to learn a unified attribute vector space, so manually labelled seed relation predicate/attribute alignments are not necessity, and we put parenthesis on the word ``seed" for relation predicate alignments and attribute alignments in Figure~\ref{fig:framework}. At least one input feature is required to train the alignment module, though.

In summary, the input features to the alignment module may be raw information such as KG structure, relation predicates, and attributes, as well as entity/relation/attribute alignments which may be created manually or automatically.

\smallvspace
\noindent \ul{Bootstrapping} { is a common strategy when limited seed alignments are available. The idea is that those aligned entity/attribute/relation produced by the EA inference module are fed back to the alignment module as training data, and this process may be iterated multiple times. Note that creating seeds takes human effort, which is expensive. Bootstrapping may help reduce human effort but is at the cost of much more computation since it iterates training multiple times.  {From the summary in Table 2, we can see that bootstrapping is very popular among the translation-based techniques but not among the GNN-based techniques. Eight out of 15 translation-based techniques exploit bootstrapping, but only one out of 17 GNN-based technique exploits bootstrapping. The reason is that the GNN-based technique is better at capturing the relationships between entities in a graph compare to the translation-based techniques. Thus, the translation-based techniques use bootstrapping to improve their capability in capturing the entity relationships. However, we believe that bootstrapping is also helpful for GNN-based techniques, which may be investigated in future work.}}

\smallvspace
\noindent 
\textbf{EA Inference module}. This module aims to predict whether a pair of entities from $\mathcal{G}_1$ and $\mathcal{G}_2$ are aligned. In practice, almost all the techniques use the following alternative aim: given a target entity $e_1$ from $\mathcal{G}_1$, the EA inference module aims to predict an entity $e_2$ from $\mathcal{G}_2$ that is aligned to $e_1$; we may call $e_1$ ($e_2$) the aligned entity or the counterpart entity of $e_2$ ($e_1$). The aligned entity may not exist if a similarity threshold is applied.

The most common approach for the inference module is \textit{nearest neighbor search} (NNS), which finds the entity from $\mathcal{G}_2$ that is the most similar to $e_1$ based on their embeddings obtained from the EA training module. Commonly used similarity measures include cosine similarity, Euclidean distance and Manhattan distance. When describing individual techniques later, we may omit the inference module if it uses this most common approach of NNS.


The NNS inference approach may incur \textit{many-to-one alignment}, where two different entities from a KG are aligned with the same entity from the other KG. To avoid it, some studies impose a one-to-one constraint.

\smallvspace
\noindent  {\textbf{Discussions}.} 
 {Table~\ref{tab:model_summary} summarizes representative EA techniques according to six key characteristics (the first row): KG structure embedding, KG structure, the way attributes used as input features, the way relation predicates used as input features, input features for the alignment module, and whether bootstrapping is used.} Column 2 captures the paradigm of the embedding module. Columns 3 to 5 describe all the raw input features. Column 6 describes the input features of the alignment module. Early techniques make use of fewer types of information. For example, MTransE~\citep{MTransE2017} only uses KG structure as features for the embedding module and seed relation triple alignments for the alignment module. Newer techniques such as AttrE~\citep{ACE2019} and MultiKE~\citep{MultiKE2019} use all the four types of raw information as input features for training modules as well as various types of features for the alignment module. As our experimental study shows (cf. Table~\ref{table-exp-1-results}), techniques using more types of input features tend to have better performance.
    \begin{table*}[htp]
    \vspace{-2mm}
        \caption{A summary of embedding-based EA techniques (``-" means \textit{not applicable})}
        \centering
        \begin{threeparttable}
        \setlength{\tabcolsep}{3.5pt}
        \begin{scriptsize}
        \begin{tabular}{lp{21mm}<{\centering}p{15mm}<{\centering}p{30mm}<{\centering}p{25mm}<{\centering}p{38mm}<{\centering}c}
        \toprule[2pt]
        \midrule
        Technique & KG structure embedding & KG structure & Relation predicates as input features & Attributes as input features & Input features for alignment module & B.\tnote{1}\\
        \midrule
        \midrule
        MTransE (2017) & TransE & Triple & - & - & Seed entity alignments; Seed relation predicate alignments & -\\
        \midrule
        IPTransE (2017) & PTransE & Path & - & - & Seed entity alignments; Seed relation predicate alignments & \checkmark\\
        \midrule
        JAPE (2017) & Modified TransE\tnote{2} & Triple & - & Data type of attribute value & Seed entity alignments; Seed relation triple alignments & -\\
        \midrule
        BootEA (2018) & Modified TransE\tnote{2} & Triple & - & - & Seed entity alignments & \checkmark\\
        \midrule
        KDCoE (2018) & TransE & Triple & - & Entity description & Seed entity alignments & \checkmark\\
        \midrule
        OTEA (2019) & TransE & Triple & - & - & Seed entity alignments & -\\
        \midrule
        SEA (2019) & TransE & Triple & - & - & Seed entity alignments & \checkmark\\
        \midrule
        NAEA (2019) & Modified TransE\tnote{2} & Neighborhood & - & - & Seed entity alignments & \checkmark\\
        \midrule
        TransEdge (2019) & TransEdge & Triple & - & - & Seed entity alignments & \checkmark\\
        \midrule
        SX19 (2019) & TransE & Triple & - & - & Seed entity alignments; Seed relation predicate alignments & -\\
        \midrule
        AKE (2019) & Modified TransE\tnote{2} & Triple & - & - & Seed entity alignments & -\\
        \midrule
        AttrE (2019) & TransE & Triple & String of relation predicate & Attribute triple as relation triple; Character sequence of attribute value; String of attribute predicate & Attribute triple as relation triple; Character sequence of attribute value; String of attribute/relation predicate & -\\
        \midrule
        MultiKE (2019) & Modified TransE\tnote{2} & Triple & String of relation predicate & Attribute triple as relation triple; String of attribute predicate/value; Entity name & Seed entity alignments; Relation/attribute predicate alignments & -\\
        \midrule
        COTSAE (2020) & TransE & Triple & - & Character sequence of attribute value/predicate & Seed entity alignments & \checkmark\\
        \midrule
        JarKA (2020) & Modified TransE\tnote{2} & Triple & - & Word embeddings of attribute value & Seed entity alignments; Seed relation/attribute predicate alignments; Seed attribute value alignments & \checkmark\\
        \midrule
        \midrule
        GCN-Align (2018) & GCN & Neighborhood & - & Attribute triple as relation triple & Seed entity alignments & -\\
        \midrule
        HMAN (2019) & GCN & Neighborhood & Bag-of-words of relation predicate & Bag-of-words of attribute value; Entity description & Seed entity alignments & -\\
        \midrule
        AVR-GCN (2019) & VR-GCN & Neighborhood & - & - & Seed entity alignments; Seed relation predicate alignments & -\\
        \midrule
        HGCN (2019) & GCN & Neighborhood & - & Entity name & Seed entity alignments, Relation predicate alignments & -\\
        \midrule
        RDGCN (2019) & DPGGNN & Neighborhood & - & Entity name & Seed entity alignments & -\\
        \midrule
        GMNN (2019) & GCN & Neighborhood & - & Entity name & Seed entity alignments & -\\
        \midrule
        MuGNN (2019) & GCN & Neighborhood & - & - & Seed entity alignments; Seed relation predicate alignments & -\\
        \midrule
        NMN (2020) & GCN & Neighborhood & - & Entity name & Seed entity alignments & -\\
        \midrule
        CG-MuAlign (2020) & CollectiveAGG & Neighborhood & - & - & Seed entity alignments & -\\
        \midrule
        CEA (2020) & GCN & Neighborhood & - & Entity name & Seed entity alignments & -\\
        \midrule
        SSP (2020) & GCN & Neighborhood & - & - & Seed entity alignments & -\\
        \midrule
        XS20 (2020) & GCN & Neighborhood & - & Entity name & Seed entity alignments & -\\
        \midrule
        \midrule
        KECG (2019) & GAT+TransE & Neighborhood & - & - & Seed entity alignments & -\\
        \midrule
        AliNet (2020) & GAT & Path & - & - &  & -\\
        \midrule
        AttrGNN (2020) & GAT & Neighborhood & - & String of attribute value; Entity name & Seed entity alignments & -\\
        \midrule
        MRAEA (2020) & GAT & Neighborhood & String of relation predicate; Direction of relation predicate & Entity name & Seed entity alignments & \checkmark\\
        \midrule
        EPEA (2020) & GAT & Neighborhood & - & String of attribute value; Entity name & Seed entity alignments & -\\
        \midrule
        \bottomrule[2pt]
        \end{tabular}
        \begin{tablenotes}
        \item[1] The column ``B." indicates whether the technique uses  bootstrapping.
        \item[2] ``Modified TransE" proposed in different papers may differ from each other.
        \end{tablenotes}
        \end{scriptsize}
        \end{threeparttable}
        \vspace{-10mm}
        \label{tab:model_summary}
    \end{table*}
    %
    
Some studies especially early ones regard whether a technique can perform EA on multilingual KGs or only on monolingual KGs as an important distinction. We argue that this is not an essential characteristic of EA techniques. The reason is that most recently proposed techniques make use of the semantic information of KGs such as the strings of relation predicates, attribute predicates and attribute values (cf. Table~\ref{tab:model_summary}), and we can perform automatic translation on the semantic information into the target language so that both KGs are in the same language, and then conduct EA on monolingual KGs such as in JarKA~\citep{JarKA2020}.  {Our experimental study (Table~\ref{table-exp-multilinguality-results}) validates this.}

\section{KG Structure Embedding Models}\label{sec:embedding}

We review two paradigms of KG structure embedding, which underlie embedding-based EA techniques.

    \subsection{Translation-based Embedding Models}\label{sec:transembedding}
    	The essence of translation-based embedding models is treating a relation in KGs as a ``translation" in a vector space between the head and the tail entities.
    	
    	\textbf{\underline{TransE}}~\citep{TransE2013}  is the first translation-based embedding model, which embeds both entities and relations into a unified (typically low-dimensional) vector space.  { {The main idea is that, if we can find a perfect suite of vector representations, i.e., \textit{embeddings}, of entities and relation predicates, then for any relation triple $(h, r, t)$, the corresponding embeddings $\boldsymbol{h}$, $\boldsymbol{r}$ and $\boldsymbol{t}$ should satisfy the vector translation operation of $\boldsymbol{h}+\boldsymbol{r} = \boldsymbol{t}$. For example, the embedding of \texttt{Victoria} plus the embedding of \texttt{capital} should equal the embedding of \texttt{Melbourne}. In other words, if we define the following function 
    	\begin{equation}
    	    f_\text{triple}(h,r,t)=||\boldsymbol{h}+\boldsymbol{r}-\boldsymbol{t}||
    	    \label{eq:trane}
    	\end{equation}
    	then ideally this function should have the value of 0 for the embeddings of all the true relation triples. Equation~\ref{eq:trane} is called the \emph{triple score function}, which measures the plausibility of a relation triple (the smaller the function value, the more likely $(h,r,t)$ form a true relation triple). In reality, the embeddings of all the entities and relation predicates are unknown and need to be learned. Further there may well not exist a suite of embeddings such that $f_\text{triple}=0$ for all the relations. Therefore, the aim of learning becomes finding a suite of embeddings that minimize the sum of $f_\text{triple}$ for all the relations in a KG. To learn effectively, \cite{TransE2013} use the strategy of negative sampling, i.e., for any $(h^\prime, r^\prime, t^\prime)$ that do not form a true relation triple, $f_\text{triple}(h^\prime, r^\prime, t^\prime)$ should be a large value, e.g., $(\texttt{Victoria}, \texttt{capital}, \texttt{dog})$; these negative samples are called \textit{corrupted relation triples} and are generated from true triples with either the head or the tail entity being replaced by a random entity. The learning process usually randomly initialize all the embeddings and then minimize a margin-loss based objective function below via gradient descent:
    	\begin{equation}
    	    \label{eq:TransELoss}	\resizebox{.92\hsize}{!}{$\mathcal{L}=\sum\limits_{(h,r,t)\in \mathcal{T}_r} \sum\limits_{(h^\prime,r,t^\prime)\in \mathcal{T}_r^\prime}\max\Big(0, \big[\gamma + f_\text{triple}(h,r, t)-f_\text{triple}(h^\prime, r^\prime, t^\prime)\big]\Big)$}
    	\end{equation}
    	where $\gamma>0$ is a margin hyper-parameter, $\mathcal{T}_r$ is the set of true relation triples, and $\mathcal{T}_r^\prime$ is a set of corrupted relation triples.}}

        After the seminal work of TransE, several variants of translation-based KG structure embedding models are proposed with improvements on the embedding space~\citep{TransH2014, TransD2015, TransG2016, ManifoldE2016} or on the triple score function~\citep{ITransF2017}. 
        Interested readers are referred to \cite{krlsurvey2017} and \cite{kgsurvey2020} for surveys on KG embedding models.
        
        \subsection{GNN-based Embedding Models}
        \label{sec:gnn-embedding}
        
        \emph{Graph neural networks} (GNNs) have yielded strong performance on graph data analysis and gained immense popularity~\citep{GNNsurvey-Wu}. There are two representative models, namely graph convolutional networks (GCNs)~\citep{GCN2016}, and graph attention networks (GAT)~\citep{GAT2017}, which will be detailed later. These two models are frequently used in recent KG embedding and EA studies because KGs are of graph structure by nature. Unlike translation-based embedding models, which treats each triple separately, GNN-based embedding models focus on aggregating information from the neighborhood of entities together with the graph structure to compute entity embeddings. The essence of GNN-based embedding models is aggregating information from the neighborhood to a target node according to rules of \emph{message passing}~\citep{message-passing2017}, i.e., the embedding information is propagated from neighbor entities to the target entity through the edges. The optimization goal of GNN-based embedding is to map entities with a similar neighborhood into embeddings close to each other in the embedding space.
        
    	\textbf{\underline{Graph Convolutional Networks (GCNs)}}~\citep{GCN2016} compute a target node's embedding as a low-dimensional vector (i.e., embedding) by aggregating the features of its neighbors in addition to itself, following the rules of message passing in graphs. Specifically, a GCN is a multi-layer GNN denoted by a function $f(\boldsymbol{X},\boldsymbol{A})$, where the inputs are feature vectors of a graph's nodes represented by a matrix $\boldsymbol{X}$ and the graph's adjacency matrix $\boldsymbol{A}$. The element $a_{ij} \in A$ indicates the connectivity between nodes $i$ and $j$, and can be viewed as the weight of the edge between the two nodes. The features of the neighbors encoded as embeddings are passed on to the target node weighted by the edge weights. This message passing process is formulated as:
    	\begin{equation}
    	\label{eq:gcn_node}
    	    \boldsymbol{Q}^{(l+1)}=\sigma (\tilde{\boldsymbol{D}}^{-\frac{1}{2}} \tilde{\boldsymbol{A}} \tilde{\boldsymbol{D}}^{-\frac{1}{2}} \boldsymbol{Q}^{(l)} \boldsymbol{W}^{(l)})
    	\end{equation}
    	where $\boldsymbol{W}^{(l)}$ is a learnable weight matrix in the $l$-th layer, $\tilde{\boldsymbol{A}}=\boldsymbol{A}+\boldsymbol{I}$ is the adjacency matrix with self-connections, $\tilde{\boldsymbol{D}}$ is a diagonal matrix of node degrees ($\tilde{\boldsymbol{D}}_{ii}=\sum_j \tilde{\boldsymbol{A}}_{ij}$), $\tilde{\boldsymbol{D}}^{-\frac{1}{2}} \tilde{\boldsymbol{A}} \tilde{\boldsymbol{D}}^{-\frac{1}{2}}$ is the normalization of $\tilde{\boldsymbol{A}}$ by node degrees, and $\boldsymbol{Q}^{(l+1)}$ is the output of the $l$-th layer, which consists of the node embeddings computed by the GCN after $l$ iterations. The input of the $l$-th layer is $\boldsymbol{Q}^{(l)}$, which in turn is the output of the previous layer. The node embeddings are usually initialized by the input matrix, i.e., $\boldsymbol{Q}^{(0)} = \boldsymbol{X}$, and the final layer produces the final node embeddings learned by the GCN. Usually $\boldsymbol{A}$ is determined by the connectivity of the graph where 1 means connected and 0 means not; $\boldsymbol{A}$ may also be determined by heuristics such as the similarity between nodes and the values may be between 0 and 1. Once $\boldsymbol{A}$ is determined, it remains unchanged during the training which means that it is not learned. 
    	
    	\textbf{\underline{Graph Attention Networks (GAT)}}~\citep{GAT2017} aggregate information from neighborhood with the \emph{attention mechanism}~\citep{attention2017} and allows for focusing on the most relevant neighbors. Conceptually, GAT is similar to GCN in the sense that they both perform message passing to compute the node embeddings; the main difference is that the edge weights of GCNs (i.e., the adjacency matrix) are not learned but those of GAT (i.e., the attentions) are. Specifically, GAT is also a multi-layer GNN denoted by a function $f(X)$ and the input $X$ is the feature vectors of nodes. The output of the $l$-th layer is computed based on the attention mechanism as follows:
    	\begin{equation}
    	    \label{eq:gat_node}
    	   \boldsymbol{q}_{i}^{(l+1)}=\sigma\Big(\sum_{j \in \mathcal{N}_{i}} \alpha_{i j} \boldsymbol{W} \boldsymbol{q}^l_{(j)}\Big)
    	\end{equation}
    	where $\boldsymbol{W}$ is a learnable weight matrix; $\boldsymbol{Q}^{(l+1)}=\boldsymbol{q}_1^{(l+1)},\boldsymbol{q}_2^{(l+1)},...,\boldsymbol{q}_n^{(l+1)}$ is the output of the $l$-th layer; $n$ is the number of the nodes; $\boldsymbol{q}_i^{(l+1)}$ is the node embedding of node $i$ computed by the GAT after $l$ iterations. The input of the $l$-th layer is $\boldsymbol{Q}^{(l)}$, which in turn is the output of the previous layer using $\boldsymbol{Q}^{(l-1)}$ as input. The node embeddings are usually initialized by the input matrix, i.e., $\boldsymbol{Q}^{(0)} = \boldsymbol{X}$, and the final layer produces the final node embeddings learned by the GAT. The attention weight $\alpha_{ij}$ is computed by a softmax normalization over attention coefficients:
    	\begin{equation}
    	\label{eq:gat_attention_weight}
    	    \alpha_{i j}=\frac{\exp(c_{ij})}{\sum_{k \in \mathcal{N}_{i}} \exp(c_{ik})}
    	\end{equation}
    	where $\mathcal{N}_{i}$ indicates the set of nodes in the neighborhood of node $i$. The attention coefficient $c_{ij}$ is the correlation between nodes, which is learned as follows:
    	\begin{equation}
    	\label{eq:gat_attention_coefficient}
    	    c_{ij} = \textsc{LeakyReLU}\big(\boldsymbol{w}^{T}[\boldsymbol{W} \boldsymbol{q}_{i}^{(l)} \| \boldsymbol{W} \boldsymbol{q}_{j}^{(l)}]\big)
    	\end{equation}
    	where the parameter vector $\boldsymbol{w}$ is used to transform the concatenation of two node embeddingss into a scalar. 
    	
    	GAT applies multi-head attention as follows: 
    	\begin{equation}
    	    \boldsymbol{q}_{i}^{(l+1)}=\big\|_{k=1}^{K} \sigma\Big(\sum_{j \in \mathcal{N}_{i}} \alpha_{i j}^{k} \boldsymbol{W}^{k} \boldsymbol{q}_{j}^{(l)}\Big)
    	\end{equation}
    	where the output is the concatenation of $K$ independent self-attentions with different normalized attention weight $\alpha_{i j}^{k}$ and weight matrix $\boldsymbol{W}^{k}$.
    	
\textbf{\underline{Variants}}. The two representative GNN models, GCN and GAT, have served as the foundation for more sophisticated models designed for various applications. Alternative symmetric matrices have been proposed to replace the adjacency matrix of GCNs (e.g., AGCN~\citep{AGCN2019} and DGCN~\citep{DGCN2018}), and various ways of computing attention have been proposed for GAT. A comprehensive discussion on GNNs is given by~\cite{GNNsurvey-Wu}.

\section{Translation-based EA Techniques}\label{sec:translationmodels}

This section reviews representative translation-based EA techniques.  { {We focus on the two key components, the embedding module determined by $f_\text{triple}$, and the alignment module determined by $f_\text{align}$. The KG embedding in translation-based EA techniques either use TransE~\citep{TransE2013} directly or its variants}}, which encodes KG structure by relation triples, paths or neighborhood. {We review the techniques that only use KG structure for their KG embedding in Section~\ref{sec:only_KG_structure} and the techniques that exploit other types of information, i.e., relation predicates and attributes for their KG embedding in Section~\ref{sec:exploit_attributes}.}

\subsection{Techniques that Only Use KG Structure}\label{sec:only_KG_structure}
\eat{
We first describe techniques whose embedding modules only use KG structural information as input features. Early techniques usually take this approach.
}

\textbf{\underline{MTransE}}~\citep{MTransE2017} is the first translation-based model for embedding-based EA. Its embedding module {uses TransE} to embed the entities and relation predicates from each KG into a different embedding space { {with part of the loss function being the same as Equation~\ref{eq:TransELoss}}}. To make these embeddings all fall into a unified space, the alignment module learns cross-KG transitions by minimizing the sum of the alignment score function for all the seed relation triple alignments as follows:
\begin{equation}
    \mathcal{L}=\sum\limits_{(tr_1, tr_2)\in\mathcal{S}_t}{f_\text{align}(tr_1, tr_2)}
\end{equation}
where $\mathcal{S}_t$ is a set of seed relation triple alignments from the $\mathcal{G}_1$ and $\mathcal{G}_2$ (essentially the combination of seed entity alignments and seed relation predicate alignments), and $f_\text{align}(tr_1,tr_2)$ is the alignment score function. {Different from the alignment score function described in Section~\ref{sec:framework}, which computes the (dis)similarity of two \textit{entities}, here the alignment score function computes the (dis)similarity of two \textit{relation triples}, $tr_1=(h_1, r_1, t_1) \in \mathcal{G}_1$ and $tr_2=(h_2, r_2, t_2) \in \mathcal{G}_2$.} To compute the alignment score, MTransE has three strategies to construct cross-KG transitions including \emph{distance-based axis calibration}, \emph{transformation vectors}, and \emph{linear transformations}. According to their experimental study, MTransE with the linear transformation strategy has the best performance.
This strategy learns a linear transformation between the entity embeddings of $\mathcal{G}_1$ and $\mathcal{G}_2$ with the following alignment score function:
\begin{align}
    f_\text{align}(tr_1, tr_2)=&||\boldsymbol{M}_{ij}^e\boldsymbol{h}_1-\boldsymbol{h}_2||+ \notag\\
    &||\boldsymbol{M}_{ij}^r\boldsymbol{r}_1-\boldsymbol{r}_2||+||\boldsymbol{M}_{ij}^e\boldsymbol{t}_1-\boldsymbol{t}_2||
\end{align}
where $\boldsymbol{M}_{ij}^e$ and $\boldsymbol{M}_{ij}^r$ are linear transformations on entity embeddings and relation predicate embeddings, respectively. Minimizing $f_\text{align}$ will minimize the distance between the transformed entities/relation predicates from $\mathcal{G}_1$ and those from $\mathcal{G}_2$, making the embeddings of the two KGs fall into the same vector space.

The inference module of MTransE uses NNS. 

\textbf{\underline{IPTransE}}~\citep{IPTransE2017} first {learns the embeddings of $\mathcal{G}_1$ and $\mathcal{G}_2$ separately in the embedding module with an extension of TransE named \emph{PTransE}~\citep{PTransE2015}). Different from TransE, PTransE can model indirectly connected entities by considering the path between them, which is composed of relation predicates that form a translation between them.}
The alignment module of IPTransE learns transitions between $\mathcal{G}_1$ and $\mathcal{G}_2$ with three different strategies based on seed entity alignments: \emph{translation-based}, \emph{linear transformation}, and \emph{parameter sharing}.

The translation-based strategy adapts the idea of  {``translation"}  to the cross-KG context, and treats alignment as a special relation predicate  $\boldsymbol{r}^{(\mathcal{E}_1 \rightarrow \mathcal{E}_2)}$ between two sets of entities, $\mathcal{E}_1$ and $\mathcal{E}_2$ from $\mathcal{G}_1$ and $\mathcal{G}_2$, respectively. The alignment score function is defined as:
\begin{equation}
	f_\text{align}(e_1,e_2)=||\boldsymbol{e}_1+\boldsymbol{r}^{(\mathcal{E}_1 \rightarrow \mathcal{E}_2)}-\boldsymbol{e}_2||
\end{equation}
where $\boldsymbol{e}_1$ and $\boldsymbol{e}_2$ are the embeddings of two entities $e_1\in \mathcal{E}_1$ and $e_2\in \mathcal{E}_2$. The objective function is a weighted sum of the loss function of PTranE and $f_\text{align}$ on seed entity alignments.

The linear transformation strategy learns a transformation matrix $\boldsymbol{M}^{(\mathcal{E}_1 \rightarrow \mathcal{E}_2)}$, which makes two aligned entities close to each other, with the alignment score function below:
\begin{equation}
f_\text{align}(e_1,e_2)=||\boldsymbol{M}^{(\mathcal{E}_1 \rightarrow \mathcal{E}_2)}\boldsymbol{e}_1-\boldsymbol{e}_2||
\end{equation}
The objective function is a weighted sum of the loss function of PTranE and $f_\text{align}$ on seed entity alignments.

The parameter sharing strategy forces $\boldsymbol{e}_1=\boldsymbol{e}_2$, which indicates that a pair of aligned entities share the same embedding, and hence applying $f_\text{align}$ on two aligned seed entities always gives 0. The objective function reduces to the loss function of PTranE. The parameter sharing strategy shows the best joint embedding learning performance among the three strategies.
	
In the training process, IPTransE adopts bootstrapping and has two strategies to add newly-aligned entities to the seeds: a hard  strategy and a soft strategy. Other techniques usually apply the hard strategy where newly-aligned entities are directly appended into the set of seed alignments, which may suffer from error propagation. In the soft strategy, reliability scores are assigned to newly aligned entities to mitigate error propagation, which correspond to the embedding distance between aligned entities. This may be implemented as a loss item added to the objective function.
	
\textbf{\underline{BootEA}}~\citep{BootEA2018} models EA as a one-to-one classification problem and the counterpart of an entity is regarded as the label of the entity. It iteratively learns the classifier via bootstrapping from both labeled data (seed entity alignments) and unlabeled data (predicated aligned entities). {The embedding module adapts the triple score function of TransE $f_\text{triple}(\cdot)$ as defined in Equation~\ref{eq:trane}} by applying $f_\text{triple}(\cdot)$ on not only true triples from $\mathcal{G}_1$ and $\mathcal{G}_2$, but also all the ``generated triples" obtained as follows: when an entity in a true triple, either head or tail, exists in the current set of aligned entities $\mathcal{S}$, replacing that entity by its aligned one in $\mathcal{S}$ generates a new triple. Note that $\mathcal{S}$ grows gradually with the iterations of bootstrapping. Specifically, the loss function for the embedding module is:
\begin{align}
    \label{eq:limit-loss}
	\mathcal{L}_{e}=&\sum\limits_{(h,r,t)\in \mathcal{T}_r}\max(0, \left[f_\text{triple}(h,r,t)-\gamma_{1}\right])+ \notag\\
	&\beta_{1} \sum\limits_{(h^\prime,r^\prime,t^\prime)\in \mathcal{T}_r^\prime}\max(0, \left[\gamma_{2}-f_\text{triple}\left(h^{\prime},r^\prime,t^{\prime}\right)\right])
\end{align}
where $\mathcal{T}_r$ includes all the true triples in $\mathcal{G}_1$ and $\mathcal{G}_2$, as well as all the generated triples described above; $\mathcal{T}^\prime_r$ contains all the corrupted triples generated by uniform negative sampling~\citep{TransE2013}.
Note that this loss function is the sum of two parts in comparison to Equation~\ref{eq:TransELoss}, which is called \textit{limit-based loss function} proposed by~\cite{limit-loss2017}; it minimizes both $f_\text{triple}(h,r,t)$ and $f_\text{triple}(h^\prime,r^\prime,t^\prime)$ by using two hyper-parameters $\gamma_{1}$ and $\gamma_{2}$ to control them directly.

The alignment module of BootEA is a one-to-one classifier, which { {is different from the aligning method in Equation \ref{margin-based-kga-loss} and}} uses a cross-entropy loss between the distribution of the entities in $\mathcal{G}_1$ vs. the distribution of the predicted class (i.e., the aligned entity) from $\mathcal{G}_2$. All the pairs of entities in $S$ are plugged into the following equation to compute the cross-entropy loss:
\begin{equation}
	\mathcal{L}_{a}=-\sum_{e_1 \in \mathcal{E}_1} \sum_{e_2 \in \mathcal{E}_2} \phi_{e_1}(e_2) \log \pi(e_2 \mid e_1)
\end{equation}
where $\phi_{e_1}(e_2)$ is a function that computes the labeling probability of $e_1$. If $e_1$ is labeled as $e_2$, the labeling distribution $\phi_{e_1}$ has all of its mass concentrated on $e_2$, i.e., $\phi_{e_1}(e_2)=1$. If $e_1$ is unlabeled, $\phi_{e_1}$ is a uniform distribution; $\pi$ is the classifier that predicts the aligned entity from $\mathcal{E}_2$ given $e_1\in\mathcal{E}_1$. 

The overall objective function of BootEA $\mathcal{L}=\mathcal{L}_e+\beta_2 \mathcal{L}_a$, where $\beta_2$ is a balancing hyperparameter.

\textbf{\underline{NAEA}}~\citep{NAEA2019} also formulates EA as a one-to-one classification problem but combines the translation-based and the GAT-based paradigms. Specifically, NAEA embeds \emph{neighbor-level information} in addition to \emph{relation-level information}. The neighbor-level information is embedded by aggregating the embeddings of the neighborhood with the attention mechanism { {as described in the GAT part of Section~\ref{sec:gnn-embedding}.}} Denote the neighbor-level representation of an entity $e$ and the neighbor-level representation of a relation predicate $r$ as $\operatorname{Ne}(e)$ and $\operatorname{Nr}(r)$, respectively. Then,  {based  {on the ``translation" idea, }} the triple score function for the neighbor-level embedding is $f_\text{triple}(h,r,t) = \|\operatorname{Ne}(h) + \operatorname{Nr}(r) - \operatorname{Ne}(t)\|$. NAEA also uses the limit-based loss trick \citep{limit-loss2017} and gets the following loss function for the neighbor-level embedding.
\begin{align}
    \mathcal{L}_{1}=&\sum_{(h,r,t) \in \mathcal{T}_r} \sum_{(h^{\prime}, r^\prime, t^\prime) \in \mathcal{T}_r^{\prime}}\max([f_\text{triple}(h,r,t)+\gamma_{1} \notag\\
    &-f_\text{triple}\left(h^{\prime},r^\prime, t^\prime\right)], 0)+ \notag\\
    &\beta_1 \sum_{(h,r,t) \in \mathcal{T}_r}\max(\left[f_\text{triple}(h,r,t)-\gamma_{2}\right], 0)
\end{align}

 {The  {relation-level information is embedded using TransE. The overall loss function for the embedding module is $\mathcal{L}_e=\beta_2 \mathcal{L}_1 + (1-\beta_2)\mathcal{L}_2$, where $\mathcal{L}_2$ is the same as Equation~\ref{eq:TransELoss} and $\beta_2$ is a hyperparameter.}}

The alignment module of NAEA is similar to BootEA, which uses a cross-entropy loss between the distribution of the entities in $\mathcal{G}_1$ and the distribution of the predicted class from $\mathcal{G}_2$ as follows.
\begin{equation}
    \mathcal{L}_a=-\sum_{e_{i} \in \mathcal{E}_1} \sum_{e_{j} \in \mathcal{E}_2} \phi_{e_1}(e_2) \log \pi\left(e_{j} \mid e_{i}\right)
\end{equation}
where $\phi_{e_1}(e_2)$ is the same as that in BootEA. The classifier $\pi\left(e_{j} \mid e_{i}\right)$ is defined as follows:
\begin{align}
    \pi\left(e_{j} \mid e_{i}\right)=& \beta_3 \smallhspace \sigma{\left(\operatorname{sim}\left(\operatorname{Ne}\left(e_{i}\right), \operatorname{Ne}\left(e_{j}\right)\right)\right)} \notag\\
    &+(1-\beta_3) \smallhspace \sigma{\left(\operatorname{sim}\left(\mathbf{e}_{i}, \mathbf{e}_{j}\right)\right)}
\end{align}
where $\operatorname{sim}(\cdot)$ is the cosine similarity and $\beta_3$ is a balancing hyperparameter.

\textbf{\underline{TransEdge}}~\citep{TransEdge2019} {addresses TransE's deficiency that its relation predicate embeddings are entity-independent}, but in reality a relation predicate embedding should depend on its context, i.e., the head and tail entities. For example, the relation predicate \texttt{director} has different meanings in two different relation triples, (\texttt{Steve Jobs, director, Apple}) and (\texttt{James Cameron, director, Avatar}).

To address this issue, TransEdge proposes an \emph{edge-centric translational embedding model}  {which  {regards the contextualized embedding of the relation predicate as the translation from the head entity to the tail entity}}. It contextualizes relation predicates as different edge embeddings, where the context of a relation predicate is specified by its head and tail entities. This is achieved by a triple score function as follows: 
\begin{equation}
    f_\text{triple}(h,r,t)=\left\| \boldsymbol{h} + \psi(\boldsymbol{h}_c, \boldsymbol{t}_c, \boldsymbol{r})-\boldsymbol{t} \right\|
\end{equation}
where $\psi(\boldsymbol{h}_c, \boldsymbol{t}_c, \boldsymbol{r})$ is the contextualized embeddings of a relation predicate, called the \textit{edge embedding}.

The paper introduces two \textit{interaction embeddings} $\boldsymbol{h}_c$ and $\boldsymbol{t}_c$ for encoding the head and tail entities' participation in the computation of the edge embeddings, respectively. The edge embeddings may be computed via two strategies, \emph{context compression} and \emph{context projection}. The first strategy, context compression, adopts muli-layer perceptrons (MLPs) to compress the embeddings of the head entity, tail entity and the relation predicate as follows:
\begin{equation}
    \resizebox{.9\hsize}{!}{$\psi(\boldsymbol{h}_c, \boldsymbol{t}_c, \boldsymbol{r}) = \textsc{MLP}\left(\smallhspace\textsc{MLP}([\boldsymbol{h}_c\|\boldsymbol{r}])\smallhspace + \smallhspace \textsc{MLP}([\boldsymbol{t}_c\|\boldsymbol{r}])\smallhspace\right)$}
\end{equation}

The other strategy, context projection, projects the relation embedding onto the hyperplane of the head and the tail entities, and compute the edge embedding as:
\begin{equation}
    \psi(\boldsymbol{h}_c, \boldsymbol{t}_c, \boldsymbol{r}) = \boldsymbol{r} - \boldsymbol{W}^\top_{(h,t)} \boldsymbol{r} \boldsymbol{W}_{(h,t)}
\end{equation}
where $\boldsymbol{W}=\textsc{MLP}([\boldsymbol{h}_c\|\boldsymbol{t}_c])$ is the normal vector of the hyperplane.

The alignment module of TransEdge uses a parameter sharing strategy to unify two different KGs, i.e., it forces a pair of aligned entities in the seed entity alignments to have the same embedding. TransEdge uses bootstrapping but newly aligned entities in each iteration are not processed with parameter sharing. To make these newly aligned entities close in the embedding space, a loss is added based on the embedding distance on the set $\mathcal{D}$ of newly aligned entities:
\begin{equation}
    \mathcal{L} = \sum_{(e_1, e_2)\in\mathcal{D}} \|\boldsymbol{e}_1-\boldsymbol{e}_2\|
\end{equation}

\noindent\textbf{\underline{Other techniques that only use KG structure}}. OTEA~\citep{OTEA2019} adapts the \emph{optimal transport} theory for EA. SEA~\citep{SEA2019} makes use of unlabeled data (unaligned entities) by adopting a cycle consistency restriction in the loss function.
%
SX19~\citep{SX2019} models \emph{multi-mappings} (i.e., many-to-many, one-to-many, or many-to-one) relations with a newly designed score function based on multiplication and complex vector space. AKE~\citep{AKE2019} first learns entity embeddings via TransE and then learns the unified vector space for $\mathcal{G}_1$ and $\mathcal{G}_2$ in an adversarial learning framework. 

\subsection{Techniques that Exploit Relation Predicates and Attributes}\label{sec:exploit_attributes}
\eat{
To obtain better \rui{what is alignment-oriented? we have removed this phrase before}alignment-oriented embeddings of KGs, some EA techniques exploit extra information from relation predicates and attributes.
}

\textbf{\underline{JAPE}}~\citep{JAPE2017} makes use of attribute triples, albeit limited to only data types of attribute values (e.g., integers or strings), in addition to relation triples.

The embedding module of JAPE has two components: \emph{structure embedding} and \emph{attribute embedding}. {The structure embeddings are obtained using TransE on $\mathcal{G}_1$ and $\mathcal{G}_2$ separately, producing two structure-based entity embedding matrices $\boldsymbol{E}_{s}^1$ and $\boldsymbol{E}_{s}^2$, respectively.} The attribute embeddings are obtained by modeling the attribute co-occurrence within a same entity or across a pair of aligned seed entities. Specifically, a word embedding (\emph{Skip-gram}~\citep{Word2Vec2013} in their paper) is computed for every data type of attribute values based on the attribute co-occurrence as described above. Then the obtained word embedding for the data type is regarded as the embedding of the attribute itself. Then we can form an attribute-based entity embedding matrix consisting of the averaged attribute embeddings of all the entities, denoted as $\boldsymbol{E}_{a}^i$ for each KG, respectively, where $i=1,2$.

After obtaining both the structure and attribute-based entity embedding matrices,
JAPE first computes cross-KG similarity $\boldsymbol{S}^{1,2}$ and inner-KG similarity for each KG (i.e., $\boldsymbol{S}^1$ and $\boldsymbol{S}^2$) based on the attribute-based entity embedding matrices:
\begin{align}
	\boldsymbol{S}^{1,2} = \boldsymbol{E}_{a}^1 {\boldsymbol{E}_{a}^2}^\top;\smallhspace
	\boldsymbol{S}^1 = \boldsymbol{E}_{a}^1 {\boldsymbol{E}_{a}^1}^\top;\smallhspace
	\boldsymbol{S}^2 = \boldsymbol{E}_{a}^2 {\boldsymbol{E}_{a}^2}^\top
\end{align}
Then, it refines the embeddings by integrating the structural information with the following loss function.
\begin{align}
	\mathcal{L}=&\left\|\boldsymbol{E}_{s}^{1}-\boldsymbol{S}^{1,2} \boldsymbol{E}_{s}^{2}\right\|_{F}^{2} + \notag\\
	&\beta \left(\left\|\boldsymbol{E}_{s}^{1}
	- \boldsymbol{S}^{1}\boldsymbol{E}_{s}^{1}\right\|_{F}^{2}
	+\left\|\boldsymbol{E}_{s}^{2}-\boldsymbol{S}^{2} \boldsymbol{E}_{s}^{2}\right\|_{F}^{2}\right)
\end{align}
where $\beta$ is a hyper-parameter that balances the importance of cross-KG similarity and inner-KG similarities.

\textbf{\underline{KDCoE}}~\citep{KDCoE2018} builds {on top of MTransE by shifting the entity embeddings by the embeddings of entity descriptions (i.e., literal descriptions for entities in KGs)}, which are treated as a type of special attribute triples where the attribute value is a literal description for the entity.

\textbf{\underline{AttrE}}~\citep{ACE2019} is the first technique that makes use of attribute values. Moreover, it is the only EA technique that needs no seed alignments.

{The embedding module of AttrE uses TransE to learn KG structure embeddings for the entities from $\mathcal{G}_1$ and $\mathcal{G}_2$.} The main novelty of AttrE is to encode the semantics of the attribute values and three methods for encoding them are proposed: \textit{averaged character embedding}, \textit{aggregated character embedding by LSTM}, and \textit{aggregated n-gram character embedding}. The aggregated n-gram character embedding has the best performance as reported in their paper, which uses the sum of n-grams of varying lengths to encode attribute values. 

Another interesting idea proposed in AttrE, { {inspired by the ``translation'' idea in Equation~\ref{eq:trane}}}, is interpreting attribute triples (in addition to relation triples) as translating operation to learn the attribute embeddings as follows:
\begin{equation}
	f_\text{triple}(e,a,v)=\left\|\boldsymbol{e}+\boldsymbol{a}-\tau(v)\right\|
\end{equation}
where $\tau(v)$ is a function implementing one of the aforementioned encoding methods on the attribute value $v$. 
Thereby the same triple score function can be used to compute the plausibility of both relation and attribute triples uniformly. It helps shift the KG structure embeddings of $\mathcal{G}_1$ and $\mathcal{G}_2$ into the same vector space by minimizing the following loss function:
\begin{equation}
\small{
	\mathcal{L}_{s}=\sum_{e \in \mathcal{E}_{1} \cup \mathcal{E}_{2}}\left[1-\operatorname{sim}\left(\boldsymbol{e}_{s}, \boldsymbol{e}_{c}\right)\right]
	}
\end{equation}\label{eq:AttrE_Loss}

where $\operatorname{sim}\left(\boldsymbol{e}_{s}, \boldsymbol{e}_{c}\right)$ is the cosine similarity between the structure embedding $\boldsymbol{e}_{s}$ and the attribute embedding $\boldsymbol{e}_{c}$ of an entity $e$. 

Besides making use of relation triples and attribute values, AttrE also aligns predicates (including both relation and attribute predicates) by exploiting the string similarity in the naming conventions of the predicates.

Finally, the inference module of AttrE predicts the aligned entity by computing the cosine similarity between the shifted structure embeddings.

\textbf{\underline{MultiKE}}~\citep{MultiKE2019} uses multi-view learning on various kinds of features. The embedding module of MultiKE divides the features of KGs into three subsets called \emph{views}: \emph{name view}, \emph{relation view}, and \emph{attribute view}. Entity embeddings are learned for each view and then combined.
	
In the name view, an entity embedding is obtained from concatenating pre-trained word/character embeddings of the tokens in the entity name. 

In the relation view, {TransE is adopted to produce embeddings but with a logistic loss function}:  
\begin{equation}
	\resizebox{.9\hsize}{!}{$\mathcal{L}_r=\sum _{(h, r, t) \in \mathcal{T}_{r} \cup \mathcal{T}_{r}^{\prime}} \log \Big(1+ \exp\big(\zeta(h,r,t) f_\text{triple}\left(h,r,t\right)\big)\Big)$}
\end{equation}
where $\zeta(h,r,t)$ is 1 if $(h,r,t)$ is a true triple, and -1 otherwise; $f_\text{triple}\left(h,r,t\right)$ is the same as in TransE. 

In the attribute view, an attribute-value matrix $[\boldsymbol{a}\|\boldsymbol{v}]$ is first formed by the concatenation of the embeddings of attribute predicates and their values, and then the triple score function is defined as the head entity minus the result of a convolution neural network (CNN) on the attribute-value matrix, formally:
\begin{equation}
\label{eq:log-transe}
	f_\text{triple}(e,a,v)=\left\|\boldsymbol{e}-\textsc{CNN}([\boldsymbol{a} \| \boldsymbol{v}])\right\|
\end{equation}
This triple score function is then used to obtain the embeddings in the attribute view by minimizing the following objective function: 
\begin{equation}
	\mathcal{L}_a=\sum_{(e, a, v) \in \mathcal{T}_{a}} \log (1+\exp (f_\text{triple}(e,a,v)))
\end{equation}
where $\mathcal{T}_{a}$ is a set of true attribute triples.

 {Next, the alignment module unifies the embedding spaces of the two KGs into the same vector space in each of the views. In the name view, the two KGs both use the same embedding scheme, i.e., pre-trained word embedding, so their embedding spaces are already unified. In the relation view and the attribute view, MultiKE performs the so-called \textit{cross-KG entity/relation/attribute identity inference} to unify the embedding spaces as follows. 

The entity identity inference is performed in both the relation and the attribute views. First, a strategy similar to BootEA~\citep{BootEA2018} is adopted to generate triples as follows: when an entity in a true triple, either head or tail, exists in the current set of aligned entities $\mathcal{S}$, replace that entity by its aligned one in $\mathcal{S}$ generates a new triple. Then the sum of the plausibility ($f_\text{triple}$) of all the generated triples is minimized in both the relation and attribute views, which update all the embeddings. The updated embeddings are then fed into the relation and attribute identity inference below.

In the relation and attribute identity inference, first a similar strategy as AttrE~\citep{ACE2019} is adopted to derive soft relation and attribute predicate alignments by string similarity. Then the relation (attribute, respectively) identity inference generates triples in the relation view (attribute view, respectively) as follows: when a relation (attribute, respectively) predicate in a true triple exists in derived relation (attribute, respectively) alignments, replace the relation (attribute, respectively) predicate with its aligned counterpart. Then the sum of the plausibility ($f_\text{triple}$) of all the generated triples is minimized in both the relation and attribute views, which update all the embeddings.

The embeddings of an entity for the three views obtained above are combined into one embedding for the entity by averaging each view or minimizing a combination loss function. Finally, the inference module uses NNS based on the similarity between the combined entity embeddings.}

\eat{
using the following loss function:
\begin{equation}
	\mathcal{L}=\sum_{i=1}^{D}\|\tilde{\boldsymbol{E}}-\boldsymbol{E}^{(i)}\|_{F}^{2}
\end{equation}
where $\tilde{\boldsymbol{E}}$ is the combined embedding matrix, $\boldsymbol{E}^{(i)}$ is the embedding matrix of the $i$-th view, and $\left\|.\right\|_{F}^{2}$ is the Frobenius norm.
}
%

 {\textbf{\underline{COTSAE}}~\citep{COTSAE2020}  alternatively trains structural and attribute embeddings, and then combines the alignment results obtained from them.}

\section{GNN-based EA Techniques}\label{sec:gnnmodels}

GNNs suit KGs' inherent graph structure so there are growing numbers of EA techniques based on GNNs recently. GNN-based EA techniques are categorized into GCN-based and GAT-based ones. They usually encode KG structure by the neighborhood of entities and many of them take attributes as input features for the embedding module because aligned entities tend to have similar neighborhood and attributes. Most GNN-based techniques use only seed entity alignments rather than other kinds of seed alignments in the training.

\subsection{GCN-based EA Techniques}


\textbf{\underline{GCN-Align}}~\citep{GCN-Align2018} is the first study on GNN-based EA. Like many GNN-based EA techniques, GCN-Align learns entity embeddings from structural information of entities. GCN-Align also exploits attribute triples by treating them as relation triples. Specifically, {GCN-Align uses two GCNs to embed the entities of $\mathcal{G}_1$ and $\mathcal{G}_2$ (one GCN for each KG) into a unified space with shared weight matrices,} described by the following equation:
\begin{equation}
	\resizebox{.9\hsize}{!}{$[\boldsymbol{H}_s^{(l+1)}\|\boldsymbol{H}_a^{(l+1)}]=\sigma(\hat{\boldsymbol{D}}^{-\frac{1}{2}}\hat{\boldsymbol{A}}\hat{\boldsymbol{D}}^{-\frac{1}{2}}[\boldsymbol{H}_s^{(l)} \boldsymbol{W}_s^{(l)}\| \boldsymbol{H}_a^{(l)} \boldsymbol{W}_a^{(l)}])$}
\end{equation}
where $\boldsymbol{H}_s^{(l)}$ and $\boldsymbol{H}_a^{(l)}$ are the matrices for the structural and attribute embeddings, respectively; $\boldsymbol{W}_s^{(l)}$ and $\boldsymbol{W}_a^{(l)}$ are the weight matrices for these two types of embeddings, which are shared by the two GCNs. {The matrix $\boldsymbol{H}^{(l)}$ in the vanilla GCN~\citep{GCN2016}  {(Equation  {\ref{eq:gcn_node})}} is replaced by a concatenation of structure and attribute embedding matrices.} Unlike GCN, GCN-Align considers various types of relation predicates in KGs when computing the element $a_{ij} \in \boldsymbol{A}$. The new adjacency matrix $\boldsymbol{A}$ is designed as follows: 
\vspace*{-1mm}
\begin{equation}
	a_{ij}\in \boldsymbol{A}=\sum\limits_{(e_i,r,e_j)\in \mathcal{T}_{r}}g_h(r) + \sum\limits_{(e_j,r,e_i)\in \mathcal{T}_{r}}g_t(r)
\end{equation}
where $a_{ij}$ is the edge weight from the $i$-th entity to the $j$-th entity. Both $(e_j,r,e_i)$ and $(e_i,r,e_j)$ are triples in a KG. The functions $g_h(r)$ and $g_t(r)$ compute the number of head entities and the number of tail entities connected by relation $r$ divided by the number of triples containing relation $r$, respectively. In this way, the adjacency matrix $\boldsymbol{A}$ helps model how the embedding information propagates across entities.

GCN-Align is trained by minimizing a margin-based loss function like Equation~\ref{margin-based-kga-loss}. Taking into account of both structure and attribute embeddings, GCN-Align defines its alignment score function as follows:
\begin{align}
	f_\text{align}(e_1,e_2)=&\beta \frac{\|\boldsymbol{h}_s(e_1)-\boldsymbol{h}_s(e_2))\|_{L_1}}{d_s}+ \notag\\
	&(1-\beta)\frac{\|\boldsymbol{h}_a(e_1)-\boldsymbol{h}_a(e_2)\|_{L_1}}{d_a}
\end{align}
where $\boldsymbol{h}_s(\cdot)$ and $\boldsymbol{h}_a(\cdot)$ are the structure embedding with dimensionality $d_s$ and attribute embedding with dimensionality $d_a$, respectively; $\beta$ is used to balance the importance of these two embeddings.

\textbf{\underline{HGCN}}~\citep{HGCN2019}  {explicitly utilizes relation representation to improve the alignment process in EA}. To incorporate the relation information, HGCN jointly learns entity and relation predicate embeddings in three stages as follows.

\textit{State 1} computes entity embeddings by a GCN variant named the \emph{Highway-GCN}~\citep{Highway-GCN2018}, which embeds entities into a unified vector space. {The layer-wise highway gates control the forward propagation on top of the vanilla GCN layer, formulated as function $T$ below}: 
\begin{equation}
    T(\boldsymbol{H}^{(l)})=\sigma \big(\boldsymbol{H}^{(l)} \boldsymbol{W}^{(l)}+\boldsymbol{b}^{(l)}\big)
\end{equation}
\vspace{-8mm}
\begin{equation}
    \resizebox{.9\hsize}{!}{$\boldsymbol{H}^{(l+1)}=T\left(\boldsymbol{H}^{(l)}\right) \odot \boldsymbol{H}^{(l+1)}+\left(\boldsymbol{1}-T\left(\boldsymbol{H}^{(l)}\right)\right) \odot \boldsymbol{H}^{(l)}$}
\end{equation}
where $\boldsymbol{H}^{(l)}$ is the output of the $l^{th}$ layer and the input of the $(l+1)^{th}$ layer, $\boldsymbol{W}^{(l)}$ and $\boldsymbol{b}^{(l)}$ are the weight matrix and bias vector, respectively; $\odot$ is element-wise multiplication.
HGCN computes entity embeddings for both KGs separately and then maps the embeddings into a unified vector space using Equation~\ref{margin-based-kga-loss}.

\textit{Stage 2}  {gets relation predicate embeddings based on their head
and tail entity representations. This stage first computes the average embeddings of all the head entities and tail entities connected to the relation predicate, respectively. The two averaged embeddings are then concatenated as the embedding of the relation predicate after a linear transformation.}

\textit{Stage 3} uses Highway-GCN again with the input being the concatenation of the entity embeddings computed in Stage 1 and the sum of all the relation predicate embeddings related to the entity. The alignment module maps the output of the Highway-GCN for the two KGs into a unified vector space with a loss similar to Equation~\ref{margin-based-kga-loss}.

\textbf{\underline{GMNN}}~\citep{GMNN2019} formulates the EA problem as graph matching between two \emph{topic entity graphs}. Every entity in a KG corresponds to a topic entity graph, which is formed by the one-hop neighbors of the entity and the corresponding relation predicates (i.e., edges). Such a graph represents the local context information of the entity. GMNN uses a graph matching model to model the similarity of two topic entity graphs, which indicates the probability of the two corresponding entities being aligned.

The graph matching model consists of four layers, including an input representation layer, a node-level matching layer, a graph-level matching layer, and a prediction layer. {The input representation layer uses a GCN to encode two topic entity graphs and obtain entity embeddings.} The node-level matching layer computes the cosine similarity between the embeddings of every pair of entities from two topic entity graphs. This layer further computes an attentive sum of entity embeddings as follows:
\vspace*{-1.5mm}
\begin{equation}
    \overline{\boldsymbol{e}}_{i}=\frac{\sum_{j=1}^{\left|\mathcal{E}_{2}\right|} \alpha_{i, j} \cdot \boldsymbol{e}_{j}}{\sum_{j=1}^{\left|\mathcal{E}_{2}\right|} \alpha_{i, j}}
    \vspace*{-2mm}
\end{equation}
where $\alpha_{i, j}$ is the cosine similarity between entity $e_i$ in a topic graph and entity $e_j$ in another topic graph. This computation is done for entities from both two topic entity graphs. The resultant weighted sum of embeddings serves as the input to the graph-level matching layer. {The graph-level matching layer runs a GCN on each topic entity graph to further propagate the local information throughout the topic entity graph. The output embeddings of the GCN is then fed to a fully-connected neural network followed by the element-wise max and mean pooling method to get the graph matching representations for each topic entity graph.} Finally the prediction layer takes the graph matching representation as input and uses a softmax regression function to predict entity alignment.

\textbf{\underline{MuGNN}}~\citep{MuGNN2019} addresses the structural heterogeneity between KGs that may result in dissimilar embeddings of the entities that should be aligned. To reconcile the heterogeneity (i.e., the difference) between the structures of $\mathcal{G}_1$ and $\mathcal{G}_2$, MuGNN uses a multi-channel GNN in the embedding module to encode a KG in multiple channels towards \textit{KG completion} and \textit{pruning exclusive entities}.

One channel of MuGNN conducts KG completion by adding the relation predicates missing from a KG using the Horn rules for each KG, e.g., $marriedTo(x; y) \land liveIn(x; z) \Rightarrow liveIn(y; z)$, as extracted by AMIE+~\citep{AMIE2015}.
The two resultant sets of rules are then transferred into each other via parameter sharing. The other channel of MuGNN prunes ``exclusive entities", i.e., those entities that only appear in one of the two KGs.

Specifically, the multi-channel GNN is formulated as follows, assuming a two-channel MuGNN:
\vspace*{-0mm}
\begin{equation}
	\resizebox{.9\hsize}{!}{$\textsc{MultiGNN}(\boldsymbol{H}^l;\boldsymbol{A}_1,\boldsymbol{A}_2)=\textsc{Pooling}(\boldsymbol{H}_1^{l+1},\boldsymbol{H}_2^{l+1})$}
\end{equation}
\vspace*{-7mm}
{\begin{equation}
    \boldsymbol{H}_i^{l+1}=\textsc{GCN}(\boldsymbol{A}_i,\boldsymbol{H}^l,\boldsymbol{W}_i), i=1,2
\end{equation}}
where,  { {similar to Equation \ref{eq:gcn_node},}} $\boldsymbol{H}^l$ is the input entity embeddings of the current layer while $\boldsymbol{H}_i^{l+1}$ is the output entity embeddings of this layer for the $i$-th channel; $\boldsymbol{A}_i$ and $\boldsymbol{W}_i$ are the adjacency matrix and learnable weight matrix in the $i$-th channel, respectively. At the end of the layer, pooling is used to combine the two channels. The adjacency matrices $A_i$ are determined by different weighting schemes with self-attention and cross-KG attentions as follows.

$\boldsymbol{A}_1$ is determined based on self-attention, where element $a_{ij}$ is the connectivity from $e_i$ to $e_j$ as follows: 
\vspace*{-3mm}
\begin{equation}
    a_{ij} = \operatorname{softmax}(c_{ij})=\frac{\exp(c_{ij})}{\sum_{e_k \in N_{e_i} \cup \{e_i\}} \exp(c_{ik})}
\end{equation}
Here, $N_{e_i}$ is the neighborhood of $e_i$, and $c_{ij}$ is the attention coefficient defined the same way as in Equation~\ref{eq:gat_attention_coefficient}.

$\boldsymbol{A}_2$ prunes exclusive entities by lowering the weight of the connectivity $a_{ij}$ between those entities if one of them is an exclusive entity, formally:
\vspace*{-1.5mm}
\begin{equation}
    \resizebox{.9\hsize}{!}{$a_{ij}\in \boldsymbol{A}_2 = \max\limits_{r_1\in \mathcal{R}_1, r_2\in \mathcal{R}_2}\boldsymbol{1}((e_i,r_1,e_j)\in \mathcal{T}_1)\operatorname{sim}(r_1,r_2)$}
\end{equation}
where, $\mathcal{R}_1$ and $\mathcal{R}_2$ are the sets of relation predicates of $\mathcal{G}_1$ and $\mathcal{G}_2$, respectively. The function $\boldsymbol{1}(\cdot) = 1$ if $(e_i,r_1,e_j) \in \mathcal{T}_1$, and 0 otherwise. The function $\operatorname{sim}(r_1,r_2)$ is the inner-product similarity between two relation predicates $r_1$ and $r_2$.

To unify embeddings of $\mathcal{G}_1$ and $\mathcal{G}_2$ from the multi-channel GNN into a same vector space, the alignment module of MuGNN utilizes a variant of Equation~\ref{margin-based-kga-loss}, which is the weighted sum of the seed entity alignments loss and the seed relation predicate alignments loss.

\textbf{\underline{NMN}}~\citep{NMN2020} aims to tackle the structural heterogeneity between KGs. To address this issue, the technique learns both the KG structure information and the neighborhood difference so that the similarities between entities can be better captured in the presence of structural heterogeneity.

To learn the KG structure information, NMN's embedding module uses a GCN with highway gates to model the KG structure information with the input of a combination of $\mathcal{G}_1$ and $\mathcal{G}_2$ to be aligned. This module is pre-trained with a margin-based loss function (cf. Equation~\ref{margin-based-kga-loss}) using seed entity alignments.

NMN then uses cross-graph matching to capture the neighborhood difference. A neighborhood sampling strategy is first used to select the more informative one-hop neighbors, based on the observation that the more often an entity and its neighbor appear in the same context, the more representative and informative the neighbor is for the entity. The cross-graph matching then compares the sampled neighborhood subgraph of an entity in the source KG with the subgraph of each candidate entity in the target KG to select an optimal aligned entity. A cross-graph vector is computed to indicate whether the entities are similar. The cross-graph matching is done by an attention mechanism.

NMN  
concatenate the entity embedding and its neighborhood representation to get the final embeddings for EA. EA is performed by measuring the Euclidean distance between entity embeddings.

\textbf{\underline{CEA}}~\citep{CEA2020} considers the dependency of alignment decisions among  entities, e.g., an entity is less likely to be an alignment target if it has already been aligned to some entity. The paper proposes a \emph{collective EA} framework. It uses structural, semantic, and string signals to capture different aspects of the similarity between entities in the source and the target KGs, which are represented by three separate similarity matrices. Specifically, {the structural similarity matrix is computed based on the embedding matrices via GCNs with cosine similarity}, the semantic similarity matrix is computed from the word embeddings, and the string similarity matrix is computed by the Levenshtein distance between the entity names.
The three matrices are further combined into a fused matrix. CEA then formulates EA as a classical stable matching problem on the fused matrix to capture interdependent EA decisions, which is solved by the \emph{deferred acceptance} algorithm~\citep{DAA2008}.

\smallvspace
\noindent \textbf{\underline{Other GCN-based EA techniques}}.
{RDGCN~\citep{RDGCN2019}, which is similar to HGCN, utilizes relation information and extends GCNs with highway gates to capture the neighborhood structural information. RDGCN differs from HGCN in that it incorporates relation information by the attentive interaction.}
AVR-GCN~\citep{AVR-GCN2019} considers multi-mappings under the GCN paradigm and learns the embeddings of entities and relation predicates simultaneously for KGs. Specifically, it first learns these embeddings via an embedding model named \emph{vectorized relational GCN}, and then uses a weight sharing mechanism to join (e.g., via concatenation or vector transformation) those embeddings into a unified vector space.
HMAN~\citep{HMAN2019} takes into account even more other types of information such as relation predicates, attribute values, and entity descriptions besides the structural information. Specifically, HMAN employs a pre-trained BERT model~\citep{BERT2019} to capture the semantic relatedness of the descriptions of two entities that cannot be measured directly.
SSP~\citep{SSP2020} uses both translation- and GNN-based paradigms. It captures local semantics from relation predicates and global structural information by a structure and semantics preserving network.
CG-MuAlign~\citep{CG-MuAlign2020} addresses structural heterogeneity by collectively aligning entities via the attention mechanism.
XS20~\citep{XuSFSY20} is another EA technique that addresses the many-to-one alignment problem in its inference module. It models EA  as a \emph{task assignment} problem and solves it by the \emph{Hungarian algorithm}~\citep{Hugarian2010}.




\subsection{GAT-based EA Techniques}
\textbf{\underline{KECG}}~\citep{KECG2019}  aims to reconcile the issue of structural heterogeneity between KGs by jointly training both a GAT-based \emph{cross-graph model} and a TransE-based \emph{knowledge embedding model}.

The cross-graph model in KECG embeds entities with two GATs on the two KGs, which encode the graph structure information. The attention mechanism in the GATs helps ignore unimportant neighbors and mitigate the issue of structural heterogeneity. {The GATs' projection matrices $\boldsymbol{W}$ (cf. Equation~\ref{eq:gat_node}) are set to diagonal matrices}, which reduces the number of  parameters to be learned and increases the model generalizability.
%

As usual, KECG uses attention mechanisms as described in Section~\ref{sec:gnn-embedding} and margin-based loss for the cross-graph model as described in Section~\ref{sec:framework}.

The knowledge embedding model in KECG uses TransE to encode the structural information in each KG separately. The overall objective function of KECG is a weighted sum of the loss functions from the cross-graph model and the knowledge embedding model.
 
\textbf{\underline{AliNet}}~\citep{AliNet2020} is based on the observation that some aligned entities from $\mathcal{G}_1$ and $\mathcal{G}_2$ do not share similar neighborhood structures. Such aligned entities may be missed by the other GNN-based EA techniques, because they rely on similar neighborhood structures for EA. 
AliNet addresses the issue by considering both direct and distant neighbors. 

AliNet learns entity embeddings by a controlled aggregation of entity neighborhood information. Without loss of generality, we describe the process for two-hop neighborhood below, although any number of hops is applicable. First, a GCN is used to aggregate the direct (i.e., one-hop) neighbors' information. Let the embedding of an entity $e_i$ at the $l$-th layer be $\boldsymbol{e}^{(l)}_{i,1}$ after one-hop neighbor aggregation. {Then for two-hop neighbors, an attention mechanism is used to indicate their contribution to the embedding of $e_i$ as follows}:
\vspace*{-0mm}
\begin{equation}
    \boldsymbol{e}_{i, 2}^{(l)}=\sigma\Big(\sum_{j \in \mathcal{N}_{2}(i) \cup\{i\}} \alpha_{i j}^{(l)} \boldsymbol{W}_{2}^{(l)} \boldsymbol{e}_{j}^{(l-1)}\Big)
\end{equation}
where $\boldsymbol{e}_{j}^{(l-1)}$ is the embedding of $e_j$ at the $(l-1)$-th layer of the GCN; $\mathcal{N}_{2}(\cdot)$ is the set of the two-hop neighbors of $e_i$; $\boldsymbol{W}_{2}^{(l)}$ is a learnable weight matrix. To retain the difference between $e_i$ and its neighbors, the attention coefficient $c_{ij}^{(l)}$ is computed using two different transformation matrices $\boldsymbol{M}^{(l)}_1$ and $\boldsymbol{M}^{(l)}_2$ for $e^{(l)}_i$ and $e^{(l)}_j$, respectively:
\begin{equation}
    c_{i j}^{(l)}=\textsc{LeakyReLU }\big[\big(\boldsymbol{M}_{1}^{(l)} \boldsymbol{e}_{i}^{(l)}\big)^{\top}\big(\boldsymbol{M}_{2}^{(l)} \boldsymbol{e}_{j}^{(l)}\big)\big]
\end{equation}

At the end of each layer of AliNet, the information from one-hop and two-hop neighbors is combined with a gating mechanism, i.e., the embedding of entity $e_i$ at the $l$-th layer is computed as follows:
\vspace*{-1.5mm}
\begin{equation}
    \boldsymbol{e}_{i}^{(l)}=g\left(\boldsymbol{e}_{i, 2}^{(l)}\right) \cdot \boldsymbol{e}_{i, 1}^{(l)}+\left(1-g\left(\boldsymbol{e}_{i, 2}^{(l)}\right)\right) \cdot \boldsymbol{e}_{i, 2}^{(l)}
\end{equation}
where $g(\cdot)$ is the gate, $g(\boldsymbol{e}_{i, 2}^{(l)})=\sigma \big(\boldsymbol{M e}_{i, 2}^{(l)}+\boldsymbol{b}\big)$, and $\boldsymbol{M}$ and $\boldsymbol{b}$ are the weight matrix and the bias.
\textbf{\underline{MRAEA}}~\citep{MRAEA2020} considers meta relation semantics including relation predicates, relation direction, and inverse relation predicates, in addition to structural information learned from the structure of relation triples. The meta relation semantics are integrated into structural embedding via \emph{meta-relation-aware embedding} and \emph{relation-aware GAT}.

To compute the meta-relation-aware embeddings (concatenation of entity and relation predicate embeddings), we first extend the set of relation triples by creating an ``inverse triple'' for each triple by replacing the original relation predicate with an ``inverse relation predicate" while keeping the same head and tail entities unchanged. Second, the entity and relation embedding components of the meta-relation-aware embeddings of the target entity are computed by averaging those of the neighbor entities, respectively.

{The relation-aware GAT} generates a structure-and-relation-aware embedding of each entity by attending the meta-relation-aware embeddings of the target entity's neighbors. Specifically, the GAT's attention coefficient $c_{ij}$, which indicates the importance of both the neighbor entity $e_j$ and the connecting relation predicate $r_k$ to the target entity $e_i$, is computed as:
\vspace*{-0mm}
\begin{equation}
    c_{ij}=\boldsymbol{w}^{T}\Big[{\boldsymbol{e}_{i}}\| \boldsymbol{e}_{j}\| \frac{1}{\left|\mathcal{M}_{ij}\right|} \sum_{r_{k} \in \mathcal{M}_{ij}} \boldsymbol{r}_{k}\Big]
\end{equation}
where the embeddings $\boldsymbol{e}_{i}$, $\boldsymbol{e}_{j}$, and $\boldsymbol{r}_{k}$ are obtained from the meta-relation-aware embeddings; $\boldsymbol{w}$ is a learnable weight vector; $\mathcal{M}_{ij}=\left\{r_{k} \mid\left(e_{i}, r_{k}, e_{j}\right) \in \mathcal{T}\right\}$ is the set of relation predicates that link from $e_i$ to $e_j$, which incorporates relation features into the attention mechanism.

{As usual, MRAEA is trained with a margin-based loss function like Equation~\ref{margin-based-kga-loss}.}

\textbf{\underline{EPEA}}~\citep{EPEA2020}  {learns embeddings of entity pairs via a \emph{pair-wise connectivity graph} (PCG) rather than embeddings for individual entities.
EPEA first generates the PCG, whose nodes are pairs of entities from $\mathcal{G}_1$ and $\mathcal{G}_2$. Given two entity pairs $(e_{1, i}, e_{2, i})$ and $(e_{1, j}, e_{2, j})$ in the PCG, an edge is added between the two entity pairs if there is a relation predicate $r_1$ connecting $e_{1, i}$ to $e_{1, j}$ in $\mathcal{G}_1$ and a relation predicate $r_2$ connecting $e_{2, i}$ to $e_{2, j}$ in $\mathcal{G}_2$. After generating the entity pairs as the nodes of the PCG, EPEA uses a CNN to encode the attributes of entity pairs into embeddings based on attribute similarity. These attribute embeddings are then fed into a GAT that further incorporates structural information and produce a score, which indicates the probability of a pair consisting of aligned entities.} This scoring function is then used as $f_\text{align}$ in Equation~\ref{margin-based-kga-loss} to train the whole model.
The inference module predicts aligned entities by performing binary classification on the scoring function value with the input being the embeddings of entity pairs.

\textbf{\underline{AttrGNN}}~\citep{AttrGNN2020} learns embeddings from both relation triples and attribute triples in a unified network. It partitions each KG into four subgraphs containing attribute triples of entity names, attribute triples of literal values, attribute triples of digital values, and the remaining triples (i.e., relation triples), respectively. For each subgraph, {entity embedding is computed based on attributes as well as the KG structure using GAT}; then a similarity matrix between $\mathcal{G}_1$ and $\mathcal{G}_2$ is computed based on the entity embedding. Finally, the four similarity matrices are averaged to yield a final similarity matrix for the inference module.

\section{Datasets and Experimental Studies}\label{sec:experiments}
We discuss the limitations of existing datasets and experimental studies, present our proposed datasets addressing the limitations, and report on a comprehensive experimental study on representative EA techniques using our datasets.

\subsection{Limitations of Existing Datasets}\label{sec:dataset-limitation}
There are several significant limitations of existing datasets, namely \textit{bijection}, \textit{lack of name variety} and, \textit{small scale}, which are detailed below.

\textbf{Bijection}. Many existing papers, including some of the benchmarking papers, have used datasets that consist of two KGs where almost every entity in one KG has one and only one aligned entity in the other KG, i.e., there is bijection between the two KGs. Such datasets have been generated from different language versions of Wikipedia (e.g., DBP15K~\citep{JAPE2017} and SRPRS$_\text{multi}$~\citep{guo2019learning}) and the application argued for such datasets is aligning two KGs in different languages, i.e., \textit{multilingual EA}. However, such application instances are infrequent in real life.

We argue that the following scenario is more common: two KGs come from different sources, e.g., a generic KG built from Wikipedia and the other from a domain-specific source such as medicine, locations, flights and music. The difference in the sources is typically not language but the coverage of knowledge, so the two KGs are complementary to each other and aligning them helps enrich them. Therefore, \textit{non-bijection} between the KGs is desired. A recent paper~\citep{zhao2020experimental} also points out that bijection is an unrealistic setting and created \textit{DBP-FB}, a dataset consisting of two KGs built from different sources, DBpedia and Freebase. It is a great step towards non-bijection datasets, but unfortunately, a big limitation of DBP-FB is that it does not contain any generic attribute (e.g., year, address, etc) triples except entity names and hence does not suffice the need of most recent EA techniques, which make heavy use of generic attributes as features. From our experiments (Table~\ref{table-exp-1-results}) we see that most recent EA techniques use generic attributes as input features which are essential for effectiveness.
%
Creating datasets coming from different sources is challenging. A recent industrial benchmark dataset MED-BBK-9K~\citep{zhang2020an} is built from different sources.
However, this dataset also has the bijection problem and the size is small: the number of unique entities covered in this dataset is less than $10,000$.

\textbf{Lack of name variety}. Unlike recently proposed datasets MED-BBK-9K and DBP-FB, most previous EA datasets are constructed from KGs with the same source. For example, DWY100K~\citep{BootEA2018} and its resampled version SRPRS$_\text{mono}$~\citep{guo2019learning} consist of KG pairs (DBpedia-Yago and DBpedia-Wikidata) with the same primary source, Wikipedia. Thus, the names of entities from two KGs may have the same surface label; such names become ``tricky features", which can be used to achieve 100\% accurate EA easily (we call this the \textit{lack of name variety problem}). To address this problem, \cite{sun2020benchmarking} remove all the entity names from DWY100K. However, this is an overkill because, in real-world settings, KGs do contain entity names as attributes but just have variety in the names of the same real-world entity. The recently proposed dataset DBP-FB have significant portion of entities ($42\%$) with different entity names and hence have good name variety due to the different data sources DBpedia and Freebase, but its lack of generic attribute triples limits its use as mentioned earlier. We need datasets that have significant amount of generic attribute triples and variety in the names. Our proposed datasets address these issues.

\textbf{Small scale}. Most existing datasets are of small (e.g., MED-BBK-9K contains 9,162 unique entities) to medium (e.g., DBP-FB contains 29,861 unique entities) sizes. Our proposed datasets contain up to 600,000 unique entities from the two KGs combined.

Table~\ref{table-dataset_comparisons} summarizes the key properties of various datasets. Our proposed benchmark DWY-NB has all the three desirable properties: non-bijection, name variety and large size. Some studies consider the property of whether the dataset is multilingual or monolingual. We do not view it as an essential property that affects the utility of the datasets, since EA techniques that utilize semantic information of KGs such as attributes can first translate it into the target language. Note that all the current multilingual datasets have the bijection problem discussed earlier.
\begin{table}[t!]
	\centering
	\caption{Dataset/Benchmark comparisons.}
	\label{table-dataset_comparisons}
	\resizebox{\columnwidth}{!}{%
    	\begin{tabular}{P{18mm}P{10mm}P{10mm}P{8mm}P{15mm}}
    		\toprule[2pt]
    		\midrule
    		\vspace{-2mm}Name & \vspace{-2mm}Non-Bijection &  
    		\vspace{-2mm}Name variety & \vspace{-2mm}Size*& \vspace{-2mm}Language \\
    		\midrule
    		\multicolumn{1}{l}{DBP15K~\citep{JAPE2017}} & No   & No & Small & Multilingual \\
    		\multicolumn{1}{l}{DWY100K~\citep{BootEA2018}} & Large   & No  & No & Monolingual \\
    		\multicolumn{1}{l}{SRPRS$_{\textbf{multi}}$~\citep{guo2019learning}} & No   & No & Small    & Multilingual \\
    		\multicolumn{1}{l}{SRPRS$_{\textbf{mono}}$~\citep{guo2019learning}} & No   & No & Small    & Monolingual \\
    		\multicolumn{1}{l}{DBP-FB~\citep{zhao2020experimental}]} & Yes    & Yes & Medium   & Monolingual \\
    		\multicolumn{1}{l}{DBP-FB~\citep{zhao2020experimental}]} & Yes    & Yes & Medium   & Monolingual \\
    		\multicolumn{1}{l}{MED-BBK-9K~\citep{zhang2020an}} & No   & Yes & Small   & Monolingual \\
    		\multicolumn{1}{l}{DWY-NB (Our Proposed)} & Yes    & Yes & Large   & Monolingual \\
    		\midrule
    		\midrule[2pt]
    		\multicolumn{5}{p{30em}}{* Dataset size represent the number of unique entities in the dataset: \textbf{Small} ($< 20,000$); \textbf{Medium} ($20,000-50,000$); \textbf{Large} ($> 50,000$)} \\
    	\end{tabular}%
    }
    \vspace{-3mm}
\end{table}%
\subsection{Limitations of Existing Experimental Studies} \label{sec:experiments_limitation}

 {Several { recent studies aim at benchmarking EA techniques. \cite{sun2020benchmarking} re-implemented 12 representative EA techniques, but the re-implementation missed important components in some techniques such as predicate alignments in AttrE~\citep{ACE2019} and MuGNN~\citep{MuGNN2019}). To avoid such problems, we use the original code of each compared technique. Another limitation of \cite{sun2020benchmarking} is that they only used bijection datasets. The study by \cite{zhao2020experimental} does not include experiments on techniques that use attribute triples such as AttrE \citep{ACE2019} and MultiKE \citep{MultiKE2019}, but as our experiments show, most recent techniques use attribute triples and they have much best performance. \cite{zhang2020an} proposes a new dataset MED-BBK-9K, but it has the limitations of bijection and small size, making the study less comprehensive.}}

\subsection{Our Proposed Benchmark DWY-NB} \label{sec:proposed_dataset}

To address the limitations of existing datasets, we propose a new benchmark called \textbf{DWY-NB} where NB stands for non-bijection. This benchmark consists of two regular-scale datasets and large-scale ones described at the end of this sub-section; each of the regular-scale datasets consists of a pair of KGs that can be used for the evaluation of EA techniques. We call the two datasets \textbf{DW-NB} and \textbf{DY-NB}. The two KGs of DW-NB are subsets of DBpedia and Wikidata~\citep{vrandecick2014wikidata}, respectively. The two KGs of DY-NB are subsets of DBpedia~\citep{auer2007dbpedia} and Yago~\citep{hoffart2013yago2}, respectively. We choose these sources as starting points because they contain rich relation and attribute triples.

Now we explain how the datasets are generated. For ease of explanation, we next use DW-NB as an example while the process for DY-NB is similar. We start from the list of aligned entities between two KGs (\textit{DBpedia and Wikidata}) from the dataset DWY100K from \citep{sun2020benchmarking}, we call this list the \textit{seed entity alignments}. This seed entity alignment contains a list of 100,000 alignments between the entities in DBpedia and Wikidata, which originally provided by the DBpedia website\footnote{\url{https://wiki.dbpedia.org/downloads-2016-10}}. We extract all those triples from DBpedia (and Wikidata, respectively) that contain the entities listed in the seed entity alignments to form a sub-graph of DBpedia (and Wikidata, respectively), and this sub-graph of DBpedia and sub-graph of Wikidata become the pair of KGs in our DW-NB dataset. 

To address the bijection problem, we randomly remove a certain percentage ($25\%$ by default) of the entities from each of the two KGs; we make sure the entities removed from one KG is different from the entities removed from the other KG so that not every entity in one KG will have an aligned entity in the other KG. As a result, by default $50\%$ of the entities in the two KGs combined do not have aligned ones. We also vary the proportion of aligned entities in our benchmark.

To address the lack of name variety problem, we add variety to the entity name attributes as follows. In every KG source (DBpedia, Wikidata, or Yago), there are often multiple attributes corresponding to the name of the entity, which we refer to as \textit{name attributes}. These multiple name attributes have different attribute values (i.e., entity names) as a result of the curation process of the KGs conducted by different humans.
For an entity with multiple name attributes, we select a name attribute that is different from that of the corresponding entity in the other KG, if any. Thus, the above procedure provides variety in entity names between the KGs in our datasets. After this process, it turns out that $36\%$ of the aligned entities have different entity names.
The statistics of DWY-NB are listed in Table~\ref{table-dataset_statistics}. To conduct scalability experiment in Experiment 5 of Section 7.4, we further generate larger versions (100K, 300K and 600K entities in each KG of the KG pair) of the dataset DW-NB in the same way as as described above. Details of all the datasets can be found at (\url{https://github.com/ruizhang-ai/EA\_for\_KG}).
%
%
%
%
%
%
%
\begin{table}[t!]
	\centering
	\caption{Statistics of our proposed benchmark DWY-NB.}
	\label{table-dataset_statistics}
	\resizebox{\columnwidth}{!}{%
		\begin{tabular}{l|r|r|r|r|r}
			\toprule[2pt]
			\midrule
			\multicolumn{1}{c|}{Subset} & \multicolumn{1}{p{3.57em}|}{Unique\newline{}entities} & \multicolumn{1}{p{3.57em}|}{Aligned\newline{}entities} & \multicolumn{1}{l|}{Predicates} & \multicolumn{1}{p{5.715em}|}{Relationship\newline{}triples} & \multicolumn{1}{p{4.215em}}{Attribute\newline{}triples} \\
			\midrule
			\midrule
			\multicolumn{6}{c}{\textbf{DW-NB}} \\
			\midrule
			DBpedia & 84,911 & \multirow{2}[0]{*}{50,000} & 545   & 203,502 & 221,591 \\
			Wikidata & 86,116 &       & 703   & 198,797 & 223,232 \\
			\midrule
			\midrule
			\multicolumn{6}{c}{\textbf{DY-NB}} \\
			\midrule
			DBpedia & 58,858 & \multirow{2}[0]{*}{15,000} & 211   & 87,676 & 173,520 \\
			Yago  & 60,228 &       & 91    & 66,546 & 186,328 \\
			\midrule
			\bottomrule[2pt]
		\end{tabular}%
	}
	\vspace*{-3mm}
\end{table}%
\subsection{ {Experiments and Results}} \label{sec:exp_settings}
We { {conduct five sets of experiments. \textit{Experiment 1}: Following the literature \citep{MTransE2017,BootEA2018,NMN2020}, the main evaluation measure for the effectiveness of EA techniques is hits@1 (or hits@k) which indicates the percentage of entities that have the correct aligned entity in the top-k predicated aligned entities. \textit{Experiment 2}: We evaluate the effect of attribute triples on the effectiveness of EA techniques as using attribute triples has been a trend of recently proposed EA techniques. 
\textit{Experiment 3}: In addition to Experiment 1, we also evaluate the effectiveness of EA techniques via direct downstream applications.
\textit{Experiment 4}: At the end of Section~\ref{sec:framework}, we argue that whether a technique is designed to conduct multilingual EA is not an essential characteristic because we can perform an automatic translation on the semantic information into the target language so that both KGs are in the same language. This experiment justifies this argument. 
\textit{Experiment 5}: We investigate how various techniques scale up with dataset sizes.}

\noindent \textbf{Compared techniques}. We compare representative techniques that provide access to their code. We do not make any changes to the code.
%
We use the parameter settings suggested in the original papers for each technique. For detailed parameter settings of each technique, readers may refer to the corresponding papers.

\noindent \textbf{Datasets}. We use the two datasets in DWY-NB (cf. Section~\ref{sec:proposed_dataset}) for the experiments.  Note that many EA techniques such as \cite{MuGNN2019,AVR-GCN2019} use manually created seed attribute/relation predicate alignments which positively impacts the performance, while some other techniques do not. To be able to isolate the effect of the factor being evaluated for a certain experiment in our experimental study (e.g., the effect of seed entity alignments proportions, the effect of attribute triples, etc.), we have aligned the predicates between the KGs in those experiments so that predicate alignments have the same effect on the performance of all the techniques  {no matter whether they take measures to align predicates. If we did not align the predicates in our data, then the techniques that do not take measures to align predicates might have poorer performance due to unaligned predicates rather than due to the effect of the factor being evaluated.}

\noindent \textbf{Environments}. We run the experiments with an Intel(R) Xeon(R) CPU E5-2650 v4 @ 2.20GHz processor, 128GB main memory, a Nvidia Tesla GPU with 32GB memory, and Ubuntu 20.4. The programming language and libraries include Python, TensorFlow, Torch, etc. depending on the language used for the original code.

}

\smallvspace
\noindent \textbf{Experiment 1: the effect of seed entity alignments.} 
 { {This experiment evaluates the accuracy of EA in terms of Hits@k while varying the amount of seed entity alignments used for training from $10\%$ to $50\%$ of the total available set of seed entity alignments (50,000 for DW-NB and 7,500 for DY-NB).} Higher hits@k means better accuracy. Table~\ref{table-exp-1-results} shows the result. The accuracy of all the techniques gets higher with more seeds, which is expected since more seeds provide more supervision.}
%

\ul{Overall, AttrE and MultiKE have much better performance than the others especially when less seed entity alignments are available.}  {\ul{This is because they make great use of various types of features such as attributes and relation predicates while the other techniques do not.}} AttrE's performance does not change while varying the amount of seeds since it does not use seed alignments, so \ul{when seed alignments are hardly available, AttrE is the clear winner}. Only when the amount of seed entity alignments reaches 50\%, NMN has slightly higher Hits@1 than AttrE.   \ul{In general, to achieve good alignment results, supervised models require seed alignments with the proportion of at least $30\%$.}

Among the GNN-based techniques, RDGCN and NMN are the top-2 in terms of Hits@1. It is worth noting that the top-2 from both translation- and GNN-based techniques exploit attribute triples, and we can see that \ul{on average, the techniques that exploit attribute triples achieve much better performance than the techniques that do not}. JAPE has poorer performance compared to other techniques that use attributes because it uses very limited information from attributes, only the data type of attribute value.

\smallvspace
 {\noindent \textbf{Experiment 2: the effect of attribute triples.} 
 {This experiment} evaluates how much benefit may be gained by exploiting attribute triples. For every EA technique, we compare the performance of a "using-attribute" version vs. a "not-using-attribute" version as follows. For a technique that exploits attribute triples by design, we get the performance of its ``not-using-attribute" version by only using relation triples (and not using attribute triples) to compute the entity embeddings. For a technique that only uses relation triples by design, we get the performance of its ``using-attribute" version by applying a naive way of exploiting attribute triples, i.e., treating the attribute triples as relation triples which means treating the attribute values as nodes in the graph. Figure~\ref{fig-exp-2-results} shows the results. The proportion of seed entity alignments used in this experiment is $30\%$ and the results on other proportions have a similar behavior. \ul{For every technique, the ``using-attribute" version outperforms the ``not-using-attribute" version, especially for those techniques that use attribute triples by design. These show that making good use of attribute triples can improve the accuracy significantly.} Among them, the gap between the two versions of AttrE is huge, because AttrE does not use seed alignments and heavily rely on attributes to train the alignment module. MultiKE uses both seed alignments and attribute triples to produce the unified embedding space, and hence has relatively smaller gap between the two versions, but still using attributes provides substantial gains. \ul{When the KGs do not contain attribute triples but seed alignments are available, MultiKE is the winner.}   {In comparison, the performance of GNN-based techniques} (e.g., HGCN, RDGCN, NMN) drops significantly (most by $50\%$) when ``not-using-attribute", as they heavily rely on entity names to initialize the node embeddings in the embedding module. In the absence of entity names, node embeddings are randomly initialized which leads to poor performance. Interestingly, the techniques that do not use attribute triples by design also gets better performance with the ``using-attribute" version, even this is by the naive way of treating attribute triples as relation triples.}
\begin{table*}[htp]
	\centering
	\caption{Experiment 1: The effect of the amount of seed entity alignments on EA accuracy in terms of Hits@k (\%)}
	\label{table-exp-1-results}
	\resizebox{0.9\textwidth}{!}{%
		\begin{tabular}{c|l|rr|rr|rr|rr|rr}
			\toprule[2pt]
			\midrule
			\multicolumn{2}{c|}{\multirow{2}[4]{*}{Technique}} & \multicolumn{2}{c|}{Seed: 10\%} & \multicolumn{2}{c|}{Seed: 20\%} & \multicolumn{2}{c|}{Seed: 30\%} & \multicolumn{2}{c|}{Seed: 40\%} & \multicolumn{2}{c}{Seed: 50\%} \\
			\cmidrule{3-12}    \multicolumn{2}{c|}{} & \multicolumn{1}{l}{Hits@1} & \multicolumn{1}{l|}{Hits@10} & \multicolumn{1}{l}{Hits@1} & \multicolumn{1}{l|}{Hits@10} & \multicolumn{1}{l}{Hits@1} & \multicolumn{1}{l|}{Hits@10} & \multicolumn{1}{l}{Hits@1} & \multicolumn{1}{l|}{Hits@10} & \multicolumn{1}{l}{Hits@1} & \multicolumn{1}{l}{Hits@10} \\
			\midrule
			\midrule
			\multicolumn{12}{c}{DW-NB} \\
			\midrule
			\multirow{8}[-2]{*}{\begin{sideways}Translation-based\end{sideways}} & MTransE & 2.82  & 10.45 & 5.42  & 18.72 & 7.88  & 25.75 & 10.42 & 31.44 & 12.98 & 36.00 \\
			& IPTransE & 5.98  & 13.45 & 7.54  & 18.78 & 12.90 & 24.61 & 16.32 & 32.86 & 23.54 & 35.98 \\
			& BootEA & 8.12  & 16.15 & 12.54 & 20.13 & 17.92 & 28.38 & 21.46 & 35.16 & 25.44 & 37.57 \\
			& TransEdge & 22.98 & 48.12 & 38.29 & 56.22 & 45.27 & 68.95 & 49.26 & 75.25 & 54.85 & 79.68 \\
			& \underline{JAPE} & 4.62  & 7.87  & 8.62  & 14.43 & 12.57 & 19.96 & 17.20 & 27.32 & 19.91 & 30.63 \\
			& \underline{MultiKE} & 80.25 & 87.58 & 82.56 & 88.92 & 84.06 & 90.05 & 84.87 & 91.24 & 85.21 & 95.06 \\
			& \underline{AttrE} & \textbf{87.98} & \textbf{95.80} & \textbf{87.98} & \textbf{95.80} & \textbf{87.98} & \textbf{95.80} & \textbf{87.98} & \textbf{95.80} & 87.98 & \textbf{95.80} \\
			\midrule
			\multirow{11}[2]{*}{\begin{sideways}GNN-based\end{sideways}} & MuGNN & 13.49 & 37.79 & 20.96 & 49.28 & 26.92 & 56.77 & 31.09 & 61.43 & 34.41 & 64.96 \\
			& AliNet & 14.58 & 31.46 & 18.55 & 35.84 & 24.34 & 50.46 & 28.39 & 55.46 & 35.31 & 58.22 \\
			& KECG  & 18.95 & 34.17 & 24.32 & 40.78 & 30.24 & 48.66 & 35.29 & 52.12 & 39.40 & 62.31 \\
			& \underline{GCN-Align} & 12.40 & 30.18 & 20.04 & 41.56 & 24.76 & 48.52 & 29.02 & 53.43 & 31.80 & 56.20 \\
			& \underline{HGCN} & 58.08 & 62.15 & 63.14 & 68.15 & 78.97 & 86.51 & 84.25 & 90.75 & 88.54 & 91.54 \\
			& \underline{GMNN} & 71.32 & 74.24 & 75.34 & 79.23 & 80.98 & 82.23 & 82.67 & 85.87 & 84.59 & 88.64 \\
			& \underline{RDGCN} & 59.22 & 62.98 & 64.22 & 68.98 & 79.02 & 87.12 & 85.34 & 90.45 & 88.21 & 93.23 \\
			& \underline{CEA}  & 50.13 & 52.31 & 63.25 & 64.12 & 80.32 & 84.21 & 84.34 & 85.54 & 86.58 & 88.34 \\
			& \underline{MRAEA} & 53.75 & 54.74 & 64.58 & 66.12 & 81.54 & 85.97 & 83.54 & 86.02 & 84.06 & 87.55 \\
			& \underline{NMN}  & 51.45 & 59.78 & 68.21 & 72.54 & 84.03 & 88.21 & 85.65 & 90.54 & \textbf{88.69} & 95.46 \\
			\midrule
			\midrule
			\multicolumn{12}{c}{DY-NB} \\
			\midrule
			\multirow{8}[-2]{*}{\begin{sideways}Translation-based\end{sideways}} & MTransE & 0.01  & 0.15  & 0.01  & 0.39  & 0.08  & 0.68  & 0.08  & 1.39  & 0.13  & 1.89 \\
			& IPTransE & 1.54  & 9.87  & 5.67  & 25.76 & 14.55 & 36.45 & 15.77 & 45.81 & 17.33 & 52.18 \\
			& BootEA & 2.15  & 14.19 & 8.47  & 38.15 & 15.77 & 48.32 & 17.22 & 57.15 & 19.24 & 58.14 \\
			& TransEdge & 22.98 & 47.50 & 37.85 & 64.85 & 48.98 & 72.15 & 58.95 & 76.54 & 62.49 & 78.54 \\
			& \underline{JAPE} & 0.70  & 1.83  & 1.57  & 3.37  & 1.40  & 3.27  & 1.37  & 1.77  & 2.37  & 4.97 \\
			& \underline{MultiKE} & 81.87 & 88.05 & 82.11 & 89.26 & 84.97 & 90.84 & 87.22 & 92.05 & 89.25 & 93.58 \\
			& \underline{AttrE} & \textbf{90.44} & \textbf{94.23} & \textbf{90.44} & \textbf{94.23} & \textbf{90.44} & \textbf{94.23} & \textbf{90.44} & \textbf{94.23} & 90.44 & 94.23 \\
			\midrule
			\multirow{11}[2]{*}{\begin{sideways}GNN-based\end{sideways}} & MuGNN & 19.16 & 51.41 & 27.40 & 62.69 & 31.60 & 68.56 & 34.73 & 71.24 & 37.15 & 74.07 \\
			& AliNet & 13.54 & 28.53 & 14.25 & 31.69 & 25.39 & 58.31 & 28.98 & 56.12 & 34.59 & 64.12 \\
			& KECG  & 11.19 & 23.65 & 14.89 & 27.25 & 20.95 & 34.48 & 22.81 & 35.44 & 24.71 & 37.15 \\
			& \underline{GCN-Align} & 8.56  & 25.09 & 17.88 & 43.88 & 24.36 & 53.43 & 31.29 & 62.44 & 33.56 & 67.88 \\
			& \underline{HGCN} & 52.54 & 64.51 & 65.87 & 77.40 & 71.14 & 85.64 & 71.45 & 85.64 & 74.54 & 87.48 \\
			& \underline{GMNN} & 62.34 & 70.34 & 64.32 & 67.34 & 75.57 & 77.47 & 78.65 & 82.65 & 82.34 & 85.62 \\
			& \underline{RDGCN} & 53.13 & 65.30 & 67.28 & 78.21 & 74.54 & 85.22 & 77.45 & 87.43 & 78.67 & 89.45 \\
			& \underline{CEA}  & 55.24 & 58.97 & 64.35 & 65.42 & 74.56 & 78.42 & 77.78 & 80.95 & 78.91 & 83.24 \\
			& \underline{MRAEA} & 52.46 & 53.20 & 60.33 & 64.54 & 73.71 & 78.52 & 74.25 & 78.66 & 76.22 & 80.15 \\
			& \underline{NMN}  & 55.74 & 64.78 & 62.54 & 70.54 & 75.87 & 80.54 & 84.55 & 88.69 & \textbf{90.78} & \textbf{94.77} \\
			\midrule
			\midrule[2pt]
			\multicolumn{12}{l}{* Techniques that use attribute triples are \ul{underlined}. The rest tables and figures follow this convention.}\\
			\multicolumn{12}{l}{* AttrE does not use any seed alignments.}
		\end{tabular}%
	}
	\vspace{-2mm}
\end{table*}%
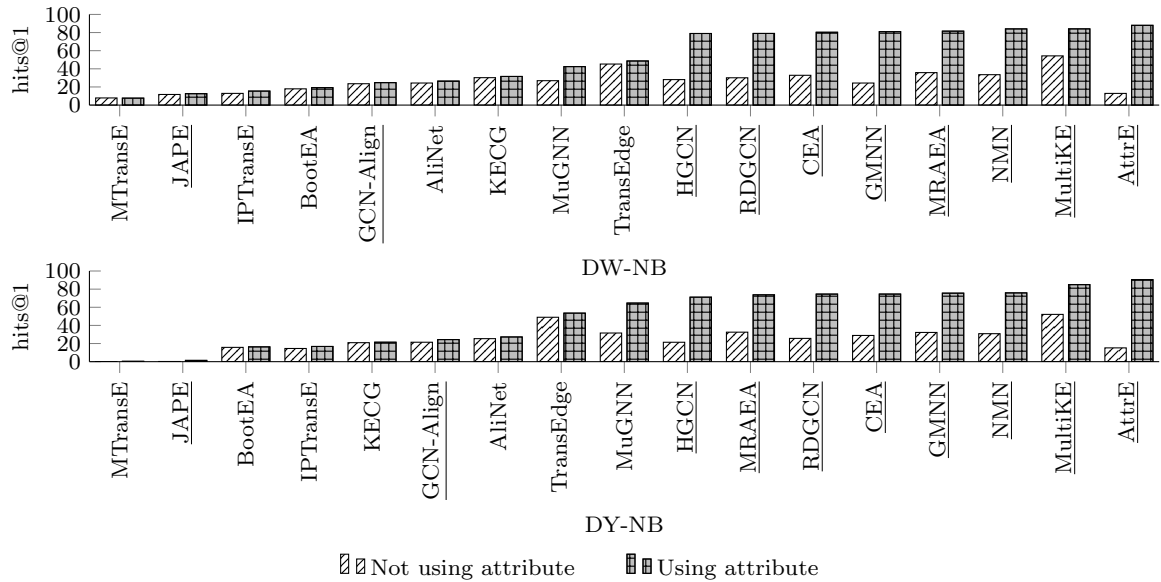
\begin{figure*}[htp]
    \centering
	\begin{tikzpicture}
	\begin{groupplot}[
	group style = {group size = 1 by 2, vertical sep=2.2cm},
	width = 0.9\textwidth,
	height = 0.16\textwidth
	]
	
	\nextgroupplot[
	enlarge x limits=0.03,
	axis x line*=bottom,
	axis y line*=left,
	ybar,
	ylabel=hits@1,
	ymin=0,
	ymax=100,
	xlabel=DW-NB,
	xtick={MTransE, \underline{JAPE}, IPTransE, BootEA, \underline{GCN-Align}, AliNet, KECG, MuGNN, TransEdge, \underline{HGCN}, \underline{RDGCN}, \underline{CEA}, \underline{GMNN}, \underline{MRAEA}, \underline{NMN}, \underline{MultiKE}, \underline{AttrE}},
	symbolic x coords={{MTransE}, {\underline{JAPE}}, {IPTransE}, {BootEA}, {\underline{GCN-Align}}, {AliNet}, {KECG}, {MuGNN}, {TransEdge}, {\underline{HGCN}}, {\underline{RDGCN}}, {\underline{CEA}}, {\underline{GMNN}}, {\underline{MRAEA}}, {\underline{NMN}}, {\underline{MultiKE}}, {\underline{AttrE}}},
	bar width=8pt,
	xticklabel style={rotate=90},
	]
	\addplot[pattern=north east lines] coordinates {(MTransE, 7.88) (\underline{JAPE}, 11.71) (IPTransE, 12.90) (BootEA, 17.92) (\underline{GCN-Align}, 23.56) (AliNet, 24.34) (KECG, 30.24) (MuGNN, 26.92) (TransEdge, 45.27) (\underline{HGCN}, 28.11) (\underline{RDGCN}, 30.12) (\underline{CEA}, 32.86) (\underline{GMNN}, 24.32) (\underline{MRAEA}, 35.80) (\underline{NMN}, 33.54) (\underline{MultiKE}, 54.29) (\underline{AttrE}, 12.88)};
	\addplot[color=black, fill=black!25, postaction={pattern=grid}] coordinates {(MTransE, 7.66) (\underline{JAPE}, 12.57) (IPTransE, 15.43) (BootEA, 19.21) (\underline{GCN-Align}, 24.76) (AliNet, 26.45) (KECG, 31.58) (MuGNN, 42.44) (TransEdge, 48.65) (\underline{HGCN}, 78.97) (\underline{RDGCN}, 79.02) (\underline{CEA}, 80.32) (\underline{GMNN}, 80.98) (\underline{MRAEA}, 81.55) (\underline{NMN}, 84.03) (\underline{MultiKE}, 84.06) (\underline{AttrE}, 87.98)};
	
	\nextgroupplot[
	enlarge x limits=0.03,
	axis x line*=bottom,
	axis y line*=left,
	ybar,
	ylabel=hits@1,
	ymin=0,
	ymax=100,
	xlabel=DY-NB,
	xtick={MTransE, \underline{JAPE}, BootEA, IPTransE, KECG, \underline{GCN-Align}, AliNet, TransEdge, MuGNN, \underline{HGCN}, \underline{MRAEA}, \underline{RDGCN}, \underline{CEA}, \underline{GMNN}, \underline{NMN}, \underline{MultiKE}, \underline{AttrE}},
	symbolic x coords={{MTransE}, {\underline{JAPE}}, {BootEA}, {IPTransE}, {KECG}, {\underline{GCN-Align}}, {AliNet}, {TransEdge}, {MuGNN}, {\underline{HGCN}}, {\underline{MRAEA}}, {\underline{RDGCN}}, {\underline{CEA}}, {\underline{GMNN}}, {\underline{NMN}}, {\underline{MultiKE}}, {\underline{AttrE}}},
	bar width=8pt,
	xticklabel style={rotate=90},
	legend style={at={($(1,3)$)},legend columns=2,fill=none,draw=none,anchor=center,align=left},
	legend to name=lg
	]
	\addplot[pattern=north east lines] coordinates {(MTransE, 0.08) (\underline{JAPE}, 0.20) (BootEA, 15.77) (IPTransE, 14.55) (KECG, 20.95) (\underline{GCN-Align}, 21.44) (AliNet, 25.39) (TransEdge, 48.98) (MuGNN, 31.60) (\underline{HGCN}, 21.41) (\underline{MRAEA}, 32.54) (\underline{RDGCN}, 25.67) (\underline{CEA}, 28.92) (\underline{GMNN}, 32.24) (\underline{NMN}, 30.87) (\underline{MultiKE}, 52.15) (\underline{AttrE}, 15.21)};
	\addplot[color=black, fill=black!25, postaction={pattern=grid}] coordinates {(MTransE, 0.39) (\underline{JAPE}, 1.40) (BootEA, 16.32) (IPTransE, 16.77) (KECG, 21.45) (\underline{GCN-Align}, 24.36) (AliNet, 27.25) (TransEdge, 53.54) (MuGNN, 64.63) (\underline{HGCN}, 71.14) (\underline{MRAEA}, 73.71) (\underline{RDGCN}, 74.54) (\underline{CEA}, 74.56) (\underline{GMNN}, 75.57) (\underline{NMN}, 75.87) (\underline{MultiKE}, 84.97) (\underline{AttrE}, 90.44)};
	
	\addlegendimage{pattern=north east lines}
	\addlegendentry{Not using attribute \hspace{5mm}}
	\addlegendimage{color=black, fill=black!25, postaction={pattern=grid}}
	\addlegendentry{Using attribute}
	
	\end{groupplot}
	\node[below] at (current bounding box.south){\pgfplotslegendfromname{lg}};
	\end{tikzpicture}
	\vspace{-2mm}
	\caption{Experiment 2: Using attributes vs. Not using attributes (sorted by performance of ``using-attribute" version).}
	\label{fig-exp-2-results}
	\vspace{-3mm}
\end{figure*}
%
%

 {
As  {a case study to understand the benefit of attribute triples intuitively, we examine the following example: \texttt{dbp:Ali\_Lohan} and \texttt{dbp:Lindsay\_Lohan} are siblings that have the same neighbors: \texttt{dbp:Michael\_Lohan} and \texttt{dbp:Dina\_lohan}, which represent the father and mother, respectively. EA techniques that only use the graph structure information cannot distinguish \texttt{dbp:Ali\_Lohan} from \texttt{dbp:Lindsay\_Lohan}, which may lead to a misalignment. EA techniques that exploit attribute triples can use the attribute triples, such as 
\texttt{(dbp:Lindsay\_Lohan, birth\_date, 1986-07-02)} 
and \texttt{(dbp:Ali\_Lohan, birth\_date, 1993-12-22)} to distinguish them. Such cases are very common in the two KGs.}}

\noindent \textbf{Experiment 3: the effect of the alignment module on KG embeddings.}  {The training in EA techniques optimize for two objectives, the \textit{KG embeddings} and the \textit{alignment of two KGs} (either jointly or alternatively), rather than merely the KG embeddings, so it might not produce the best KG embeddings.  {This experiment evaluates how the quality of the KG embeddings obtained from EA techniques are affected compared to the KG embeddings obtained from pure KG embedding techniques (TransE for translation-based and GCN for GNN-based techniques) via downstream applications of KGs.} Following previous studies in EA techniques \citep{sun2020benchmarking,zhao2020experimental}, we use two common downstream tasks for this purpose: \textit{link prediction} for translation-based techniques and \textit{node classification} for GNN-based techniques, detailed as follows. The link prediction task aims to predict $t$ given $h$ and $r$ of a relation triple. Specifically, first a relation triple is corrupted by replacing its tail entity with all the entities in the dataset. Then, the corrupted triples are ranked in ascending order by the plausibility score computed as $\boldsymbol{h} + \boldsymbol{r} - \boldsymbol{t}$. Since true triples (i.e., the triples in a KG) are expected to have smaller plausibility scores and rank higher in the list than the corrupted ones, hits@10 (whether the true triples are in the top-10) is used as the measure for the link prediction task. The node classification task aims to classify nodes and determine their labels. Given the embedding of a node, a simple classifier SVM~\citep{cortes1995support} with two-fold cross-validation is trained to predict the entity type (e.g., \texttt{person, organization}, etc.) of the node. 
 {Table~\ref{table-exp-4-results} shows} the accuracy of downstream applications on DWY-NB with $10\%$, $30\%$, and $50\%$ of seed entity alignments. The accuracy increases with the amount of seed alignments but not significantly.
\begin{table}[tp!]
	\begin{center}
		\caption{Experiment 3: effects on downstream tasks.}\vspace*{-2mm}
		\label{table-exp-4-results}
		\resizebox{\columnwidth}{!}{%
    		\begin{tabular}{@{}lllllll}
                \toprule[2pt]
                \midrule
                \multicolumn{1}{@{}c|}{\multirow{2}[4]{*}{Technique}} & \multicolumn{3}{c|}{DBP-WD (seed)} & \multicolumn{3}{c}{DBP-YAGO (seed)} \\
            \cmidrule{2-7}    \multicolumn{1}{c|}{} & \multicolumn{1}{c}{(10\%)} & \multicolumn{1}{c}{(30\%)} & \multicolumn{1}{c|}{(50\%)} & \multicolumn{1}{c}{(10\%)} & \multicolumn{1}{c}{(30\%)} & \multicolumn{1}{c}{(50\%)} \\
                \midrule[1.5pt]
                \multicolumn{7}{c}{Link Prediction (Evaluating Translation-based Models)} \\
                \midrule[1.5pt]
                \multicolumn{1}{l|}{\ul{MultiKE}} & \multicolumn{1}{r}{\textbf{88.76}} & \multicolumn{1}{r}{\textbf{88.98}} & \multicolumn{1}{r|}{\textbf{89.52}} & \multicolumn{1}{r}{98.62} & \multicolumn{1}{r}{\textbf{98.87}} & \multicolumn{1}{r}{98.07} \\
                \multicolumn{1}{l|}{\ul{AttrE}} & \multicolumn{1}{r}{88.50} & \multicolumn{1}{r}{88.50} & \multicolumn{1}{r|}{88.50} & \multicolumn{1}{r}{\textbf{98.75}} & \multicolumn{1}{r}{98.75} & \multicolumn{1}{r}{\textbf{98.75}} \\
                \multicolumn{1}{l|}{TransE*} & \multicolumn{1}{r}{87.45} & \multicolumn{1}{r}{87.45} & \multicolumn{1}{r|}{87.45} & \multicolumn{1}{r}{98.42} & \multicolumn{1}{r}{98.42} & \multicolumn{1}{r}{98.42} \\
                \multicolumn{1}{l|}{TransEdge} & \multicolumn{1}{r}{85.27} & \multicolumn{1}{r}{85.71} & \multicolumn{1}{r|}{86.40} & \multicolumn{1}{r}{93.24} & \multicolumn{1}{r}{93.54} & \multicolumn{1}{r}{93.76} \\
                \multicolumn{1}{l|}{\ul{JAPE}} & \multicolumn{1}{r}{83.24} & \multicolumn{1}{r}{83.71} & \multicolumn{1}{r|}{83.09} & \multicolumn{1}{r}{75.03} & \multicolumn{1}{r}{75.32} & \multicolumn{1}{r}{75.66} \\
                \multicolumn{1}{l|}{IPTransE} & \multicolumn{1}{r}{81.06} & \multicolumn{1}{r}{81.23} & \multicolumn{1}{r|}{81.78} & \multicolumn{1}{r}{93.10} & \multicolumn{1}{r}{93.55} & \multicolumn{1}{r}{93.91} \\
                \multicolumn{1}{l|}{BootEA} & \multicolumn{1}{r}{80.41} & \multicolumn{1}{r}{80.90} & \multicolumn{1}{r|}{81.66} & \multicolumn{1}{r}{94.11} & \multicolumn{1}{r}{94.54} & \multicolumn{1}{r}{94.85} \\
                \multicolumn{1}{l|}{MTransE} & \multicolumn{1}{r}{80.10} & \multicolumn{1}{r}{80.33} & \multicolumn{1}{r|}{80.69} & \multicolumn{1}{r}{93.81} & \multicolumn{1}{r}{94.31} & \multicolumn{1}{r}{94.74} \\
                \midrule[1.5pt]
                \multicolumn{7}{c}{Node Classification (Evaluating GNN-based Models)} \\
                \midrule[1.5pt]
                \multicolumn{1}{l|}{GCN*} & \multicolumn{1}{r}{\textbf{64.93}} & \multicolumn{1}{r}{\textbf{64.93}} & \multicolumn{1}{r|}{\textbf{64.93}} & \multicolumn{1}{r}{\textbf{68.21}} & \multicolumn{1}{r}{\textbf{68.21}} & \multicolumn{1}{r}{\textbf{68.21}} \\
                \multicolumn{1}{l|}{\ul{NMN}} & \multicolumn{1}{r}{62.25} & \multicolumn{1}{r}{62.74} & \multicolumn{1}{r|}{62.85} & \multicolumn{1}{r}{66.46} & \multicolumn{1}{r}{66.57} & \multicolumn{1}{r}{66.79} \\
                \multicolumn{1}{l|}{\ul{CEA}} & \multicolumn{1}{r}{60.06} & \multicolumn{1}{r}{60.24} & \multicolumn{1}{r|}{60.39} & \multicolumn{1}{r}{65.95} & \multicolumn{1}{r}{66.31} & \multicolumn{1}{r}{66.66} \\
                \multicolumn{1}{l|}{\ul{MRAEA}} & \multicolumn{1}{r}{57.95} & \multicolumn{1}{r}{58.34} & \multicolumn{1}{r|}{58.56} & \multicolumn{1}{r}{65.77} & \multicolumn{1}{r}{65.87} & \multicolumn{1}{r}{66.37} \\
                \multicolumn{1}{l|}{\ul{GCN-Align}} & \multicolumn{1}{r}{54.05} & \multicolumn{1}{r}{54.52} & \multicolumn{1}{r|}{54.93} & \multicolumn{1}{r}{61.54} & \multicolumn{1}{r}{62.03} & \multicolumn{1}{r}{62.37} \\
                \multicolumn{1}{l|}{\ul{HGCN}} & \multicolumn{1}{r}{53.93} & \multicolumn{1}{r}{54.11} & \multicolumn{1}{r|}{54.59} & \multicolumn{1}{r}{65.79} & \multicolumn{1}{r}{66.24} & \multicolumn{1}{r}{66.48} \\
                \multicolumn{1}{l|}{\ul{GMNN}} & \multicolumn{1}{r}{52.21} & \multicolumn{1}{r}{52.36} & \multicolumn{1}{r|}{52.67} & \multicolumn{1}{r}{67.63} & \multicolumn{1}{r}{67.87} & \multicolumn{1}{r}{67.95} \\
                \multicolumn{1}{l|}{MuGNN} & \multicolumn{1}{r}{51.63} & \multicolumn{1}{r}{51.96} & \multicolumn{1}{r|}{52.46} & \multicolumn{1}{r}{43.16} & \multicolumn{1}{r}{43.45} & \multicolumn{1}{r}{43.58} \\
                \multicolumn{1}{l|}{\ul{RDGCN}} & \multicolumn{1}{r}{51.31} & \multicolumn{1}{r}{51.46} & \multicolumn{1}{r|}{51.86} & \multicolumn{1}{r}{56.80} & \multicolumn{1}{r}{57.00} & \multicolumn{1}{r}{57.28} \\
                \multicolumn{1}{l|}{KECG} & \multicolumn{1}{r}{44.83} & \multicolumn{1}{r}{45.19} & \multicolumn{1}{r|}{45.50} & \multicolumn{1}{r}{57.75} & \multicolumn{1}{r}{58.12} & \multicolumn{1}{r}{58.45} \\
                \multicolumn{1}{l|}{Alinet} & \multicolumn{1}{r}{42.68} & \multicolumn{1}{r}{42.98} & \multicolumn{1}{r|}{43.16} & \multicolumn{1}{r}{37.96} & \multicolumn{1}{r}{38.45} & \multicolumn{1}{r}{38.72} \\
                \midrule
                \midrule[2pt]
                \multicolumn{7}{l}{* baseline} \\
            \end{tabular}%
		}
	\end{center}
	\vspace{-8mm}
\end{table}

Translation-based EA techniques are compared against TransE, a pure KG embedding technique. \ul{MultiKE and AttrE have higher link prediction accuracy than TransE while the others do not}.  {This is because MultiKE and AttrE make great use of various types of information including attribute triples as input features, which improve the quality of KG embeddings.}

GNN-based EA techniques are compared against GCN, a pure KG embedding technique. \ul{All of the GNN-based techniques have lower node classification accuracy than GCN}; the best one is NMN (about 2\% lower than GCN). The techniques that use attribute triples achieve better accuracy than those that do not. 

 \ul{In summary, the KG embeddings obtained from EA techniques may have slightly better or worse performance in downstream tasks depending on the paradigm of KG structure embeddings,} details provided in previous paragraphs.}




\smallvspace
 {\noindent \textbf{Experiment 4:}  {\textbf{Multilinguality}.} This experiment evaluates how various techniques perform on multilingual KGs with the approach of first translating into the same language. Following previous studies \citep{zhao2020experimental} we use the multilingual dataset SRPRS$_\text{multi}$~\citep{guo2019learning}, which contains two KG pairs EN-DE and EN-FR. To translate the attribute triples into English, we use a popular open-source translation tool \textit{Fairseq}~\citep{ott2019fairseq}. For each technique we run a version ``not using attributes" (the original techniques) and a version ``using attributes" (the translation approach). The results are shown in Table~\ref{table-exp-multilinguality-results}. All the techniques have significant improvement by using attributes via the translation approach, including the techniques that can perform multilingual EA by design (mostly all the GNN-based techniques). AttrE and MultiKE are not designed for multilingual EA, but via the translation approach both have comparable performance to the techniques designed for multilingual EA. \ul{These validate our argument that techniques designed for monolingual EA can perform multilingual EA well by exploiting semantic information (such as attributes) and automatic translation.}}

\smallvspace
\noindent \textbf{Experiment 5:}  {\textbf{Scalability}.}  This experiment evaluates how various techniques perform as data sizes grow. We use the same way described in Section~\ref{sec:proposed_dataset} to create EA datasets with varying numbers of entities 100K, 300K, and 600K in each KG of the KG pair.
The sources of the pair of KGs are DBpedia and Wikidata, so we call them \textbf{DW-NB-100K}, \textbf{DW-NB-300K} and \textbf{DW-NB-600K}, respectively. We have addressed the bijection and name variety problems in them such that the numbers of seed entity alignments are around 50K, 150K, and 300K, respectively. For each dataset, we use $30\%$ of the aligned entities for training. We focus our experiments on four representative techniques, AttrE, MultiKE, NMN and MRAEA, the top-2 from translation- and GNN-based techniques on the DW-NB dataset.
\begin{table}[t!]
	\begin{center}
		\caption{Effects of multilingual KGs}
		\label{table-exp-multilinguality-results}
		\resizebox{.9\columnwidth}{!}{%
    		\begin{tabular}{l|cc|cc}
                \toprule[2pt]
                \midrule
                \multicolumn{1}{c|}{\multirow{3}[6]{*}{Technique}} & \multicolumn{4}{c}{Hits@1} \\
                \cmidrule{2-5}          & \multicolumn{2}{c|}{Not using attributes} & \multicolumn{2}{c}{Using attribute} \\
                \cmidrule{2-5}          & EN-DE & EN-FR & EN-DE & EN-FR \\
                \midrule
                MTransE & 14.51 & 8.58  & 21.78 & 13.31 \\
                IPTransE & 8.09  & 9.45  & 12.80 & 14.46 \\
                BootEA & 24.67 & 35.20 & 36.66 & 51.50 \\
                TransEdge & 27.53 & 37.81 & 40.11 & 55.21 \\
                MuGNN & 15.61 & 19.44 & 23.22 & 28.81 \\
                Alinet & 14.07 & 18.36 & 20.82 & 27.30 \\
                KECG  & 20.90 & 20.34 & 31.24 & 30.28 \\
                JAPE  & 15.86 & 19.90 & 24.00 & 29.44 \\
                MultiKE & 46.53 & 41.78 & 67.56 & 60.56 \\
                AttrE & 14.55 & 12.83 & 64.74 & 56.79 \\
                GCN-Align & 21.18 & 30.86 & 31.10 & 45.38 \\
                HGCN  & 46.78 & 38.25 & 67.76 & 55.91 \\
                GMNN  & 46.53 & 38.09 & 67.63 & 55.28 \\
                RDGCN & 46.06 & 39.77 & 66.72 & 57.95 \\
                CEA   & 46.72 & 43.97 & 67.83 & 63.83 \\
                MRAEA & 47.67 & 43.25 & 69.13 & 62.67 \\
                NMN   & 48.08 & 42.96 & 69.47 & 62.72 \\
                \midrule
                \bottomrule[2pt]
            \end{tabular}%
		}
	\end{center}
	\vspace{-8mm}
\end{table}

\noindent
\underline{\textit{Theoretical analysis}}. 
For simplicity, suppose both KGs have a similar number of entities $N$. Let $M$ denote the total number of triples  (i.e., the number of edges) in the two KGs; then $M$ is $N^2$ in the worst case but $M << N^2$ in practice as the graph is sparse.

The inference module is typically via NNS or similar operations, which has the time/space cost of $\mathcal{O}(N)$ for each entity; for for aligning all the entities, the time cost is $\mathcal{O}(N^2)$ and space cost is $\mathcal{O}(N)$.

The training module includes an embedding module and an alignment module. The time/space cost for the embedding module is $\mathcal{O}(M)$ as it iterates through all the triples in the two KGs as well as a fixed number of negative samples per positive sample (i.e. per triple). The time/space cost of the alignment module depends on the algorithm. Most techniques iterate through the seed entity alignments and optionally a fixed number of negative alignment samples per seed entity alignment, so the cost is $\mathcal{O}(|\mathcal{S}|)$, where $|\mathcal{S}|$ is the number of seed entity alignments. AttrE is a special case as it does not use seeds; its cost is $O(N)$ according to Equation~\ref{eq:AttrE_Loss}. The space cost of alignment is $O(N)$. The space cost for the training is $\mathcal{O}(N)$. 

\textit{Note.} Although translation- and GNN-based techniques have the same asymptotic training cost, their practical GPU memory usage differs greatly due to different training mechanisms. Each training iteration of translation-based techniques typically requires computing the translation function (Equation~\ref{eq:TransELoss}) or its variant, which only involves several triples. The two KGs and the embeddings are stored in the CPU memory. A machine learning framework such as TensorFlow only loads the triples needed for each training iteration, namely mini-batch, from the CPU memory to the GPU memory. We can control the mini-batch size to be as small as only several triples, so the required GPU memory for translation-based techniques is very small. In comparison, GNN-based techniques usually use message passing to compute the embedding of graph nodes and edges, and the message passing procedure in TensorFlow loads the whole graph into the GPU memory, which is $\mathcal{O}(M)$, a large number. GPU memory is a bottleneck for running machine learning algorithms. This makes translation-based techniques more scalable than GNN-based ones in terms of GPU memory requirement.

\noindent
\underline{\textit{Experimental results}}.
Table~\ref{table-exp-runtime-results} shows the running time and GPU memory usage of the training module of four representative techniques as data sizes grow. 
\begin{table}[t!]
	\begin{center}
		\caption{Running time and GPU memory vs. dataset sizes}\vspace*{-2mm}
		\label{table-exp-runtime-results}
		\resizebox{\columnwidth}{!}{%
    		\begin{tabular}{l|rrr|rrr}
            \toprule[2pt]
            \midrule
            \multicolumn{1}{c|}{\multirow{2}[4]{*}{Technique}} & \multicolumn{3}{p{10.715em}|}{Running time (days)} & \multicolumn{3}{c}{GPU memory usage (GB)} \\
        \cmidrule{2-7}          & \multicolumn{1}{c}{100K} & \multicolumn{1}{c}{300K} & \multicolumn{1}{c|}{600K} & \multicolumn{1}{c}{100K} & \multicolumn{1}{c}{300K} & \multicolumn{1}{c}{600K} \\
            \midrule
            AttrE &            2.2  &     5.6  &    10.9   & 4.5 & 4.5 & 4.5 \\
            MultiKE &            2.2  &     6.0  &  11.5   & 7.3 & 7.3 & 7.3 \\
            NMN   &            2.8  &    6.1   &  N/A  & 11.2 &   28.6    & N/A \\
            MRAEA &            2.5  &     5.7  &  N/A  & 11.5 & 28.6 & N/A \\
            \midrule
            \bottomrule[2pt]
            \end{tabular}%
		}
	\end{center}
\end{table}
\begin{table}[t!]
\vspace{-5mm}
	\begin{center}
		\caption{Accuracy vs. dataset sizes}\vspace*{-2mm}
		\label{table-exp-scalability-results}
		\resizebox{\columnwidth}{!}{%
    		\begin{tabular}{l|rr|rr|rr}
                \toprule[2pt]
                \midrule
                \multicolumn{1}{c|}{\multirow{2}[4]{*}{Technique}} & \multicolumn{2}{c|}{100K} & \multicolumn{2}{c|}{300K} & \multicolumn{2}{c}{600K} \\
                \cmidrule{2-7}          & \multicolumn{1}{l}{Hits@1} & \multicolumn{1}{l|}{Hits@10} & \multicolumn{1}{l}{Hits@1} & \multicolumn{1}{l|}{Hits@10} & \multicolumn{1}{l}{Hits@1} & \multicolumn{1}{l}{Hits@10} \\
                \midrule
                AttrE & 75.59 & 80.30 & 70.59 & 74.95 &   61.22    & 64.98 \\
                MultiKE & 75.86 & 79.86 & 69.56 & 72.48 &    61.42   & 65.53 \\
                NMN   & 75.18 & 78.68 &   70.61    &    73.28   & N/A   & N/A \\
                MRAEA & 71.70 & 77.00 & 69.18 & 72.33 & N/A   & N/A \\
                \midrule
                \bottomrule[2pt]
            \end{tabular}%
		}
	\end{center}
	\vspace*{-7mm}
\end{table}
The running time of all the tested techniques is similar for the same dataset size. Translation-based techniques, AttrE and MultiKE, have constant GPU memory usage, because it is determined by the size of mini-batches. MultiKE uses more GPU memory than AttrE because MultiKE is more complicated and computes more things. As the data size grows from 100K to 300K, the GPU memory usage of GNN-based techniques grows more than translation-based ones because it is $\mathcal{O}(M)$. We are not able to run the two GNN-based techniques on the 600K dataset as the GPU memory on our server (32GB) is not enough. \ul{In summary, translation-based techniques are more scalable than GNN-based ones in practice. We observe that the growth of running time and GPU memory usage of GNN-based techniques is sub-linear, which is due to great sparsity of the graphs.} The inference modules of all the techniques take two to three hours, much less than training.


Table~\ref{table-exp-scalability-results} shows the accuracy of the techniques as the dataset size grows. \ul{The accuracy of all the techniques degrades a little as the number of entities increases.} This is because for the same entity in a KG, there are more entities in the other KG similar to it in the case of larger datasets, making it harder to predict the aligned entity correctly.


\noindent \underline{\textit{Alignment on large-scale KGs}}. An interesting question is whether the algorithms can scale up to large-scale KGs such as Wikidata (95 million entities) with Freebase (86 million entities). In what follows, we estimate the running time for aligning large-scale KGs using the CPU and GPU servers available to us, which has 128GB CPU main memory and 32GB GPU memory. It turns out to be over 2000 days. We then give an idea of how one may do it within days if large numbers of GPU servers are available.

As we discussed earlier, GNN-based techniques require loading the whole graph into the GPU memory. The full Wikidata  or Freebase KGs would require hundreds of GB memory to even store the embeddings of the entities, so we will not be able to do the experiment using GNN-based techniques. 
For translation-based techniques, the GPU memory used for training is constant. The total training time mainly consists of the time for the embedding module and the aligning module. The embedding time of translation-based models is proportional to M, the number of triples in the two KGs. The alignment time for techniques that use seed entity alignments such as MultiKE is proportional to $|S|$, the number of seed entity alignments. Our experiment has followed previous work to set $|S|$ as a certain percentage of the total number of entities N, so the alignment time is also proportional to N. AttrE does not use seed alignments, and its alignment time is proportional to N, the number of entities in the two KGs. The training time for AttrE and MultiKE are 2.2 days for the 100K entities dataset as shown in Table 8. We have recorded the training time for the embedding and alignment modules, respectively, as shown in Table~\ref{training_time_estimation}. Take AttrE as an example, the embedding time is 1.76 days and the alignment time is 0.44 days, which add up to 2.2 days. 

\begin{table}[t!]
	\begin{center}
		\caption{Estimate of the training time (days) for aligning Wikidata and Freebase}
		\label{training_time_estimation}
		\resizebox{\columnwidth}{!}{%
    		\begin{tabular}{l|c c c|c}
                \toprule[2pt]
                \midrule
                \multicolumn{1}{c|}{\multirow{2}[4]{*}{Technique}} & \multicolumn{3}{c|}{DW-NB-100K} & \multicolumn{1}{c}{Wikidata\&Freebase}\\
                \cmidrule{2-5}
                & Embedding & Alignment & Total & Training time\\
                \midrule
                AttrE & 1.76 & 0.44 & 2.2 & $1.76\times1120M/0.79M + 
 0.44\times184M/0.2M = 2,900$\\
                MultiKE & 1.88 & 0.32 & 2.2	& $1.88\times1120M/0.79M + 
0.32\times184M/0.2M = 2,960$ \\
                \midrule
                \bottomrule[2pt]
            \end{tabular}%
		}
	\end{center}
	\vspace*{-5mm}
\end{table}

There are totally 1,120M triples in Wikidata (782M triples) and Freebase (338M triples) together while there are $0.79$M triples in the DW-NB-100K dataset, so the total embedding time for aligning Wiki-Freebase is $1.76 \times 1,120M /0.79M=2495.2$ days for AttrE. There are totally 184M entities in Wikidata (98M entities) and Freebase (86M entities) while there are 0.2M entities in 100K dataset, so the alignment module running time for AttrE is $0.44\times184M/0.2M = 404.8$ days. The total training time for AttrE is $1.76\times1120M/0.79M + 0.44\times184M/0.2M = 2,900$ days. We can estimate the training time for MultiKE similarly as $1.88\times1120M/0.79M + 0.32\times184M/0.2M = 2,960$ days. Therefore, using the academic resources available to us, it is impossible to run this experiment in practice. Large companies may have thousands of GPUs and run machine learning tasks in parallel. A straightforward approach to parallelize the algorithm is like the block nested-loop join algorithm. We may divide the two KGs into $i$ and $j$ blocks, and then align each of the $i$ blocks with each of the $j$ blocks. Every block pair to be aligned can be computed on a GPU server. This way, we may be able to perform the training of the alignment within days if we use thousands of GPU servers in parallel. Of course, there are details to be worked out in terms of how to coordinate all the nodes and deal with the effect that now the alignment is computed locally rather than globally.  

Another question is the accuracy of the algorithms at large-scale. The accuracy depends on a number of factors including: the alignment algorithms, the amount of seed entity alignments, the nature of the datasets, the embedding size, etc. We have seen from Table 5 that the accuracy increases with the amount of seed alignments, and can be above 95\% for some of the cases. One approach to increase accuracy is enlarging the embedding size. We have conducted experiments for the effect of embedding size on the accuracy of the algorithms using the 100K DBpedia-Wikidata dataset, and the results are shown in Table~\ref{embedding_size_results}. It shows how the accuracy of the different algorithms increases as the embedding size grows from 128 dimensions to 512 dimensions. The accuracy increase plateaus after the number of dimensions is large enough.

\begin{table}[t!]
	\begin{center}
		\caption{Experiments on embedding size (DW-NB-100K)}
		\label{embedding_size_results}
		\resizebox{\columnwidth}{!}{%
    		\begin{tabular}{l|rr|rr|rr}
                \toprule[2pt]
                \midrule
                \multicolumn{1}{c|}{\multirow{2}[4]{*}{Technique}} & \multicolumn{2}{c|}{128} & \multicolumn{2}{c|}{256} & \multicolumn{2}{c}{512} \\
                \cmidrule{2-7}          & \multicolumn{1}{l}{Hits@1} & \multicolumn{1}{l|}{Hits@10} & \multicolumn{1}{l}{Hits@1} & \multicolumn{1}{l|}{Hits@10} & \multicolumn{1}{l}{Hits@1} & \multicolumn{1}{l}{Hits@10} \\
                \midrule
                AttrE & 52.17 & 59.31 & 76.23 & 81.17 & 76.98 & 83.22\\
                MultiKE & 51.88 & 57.89 & 76.55 & 80.37 & 77.11 & 80.92\\
                NMN & 54.23 & 56.23 & 75.76 & 79.65 & 76.24 & 80.32\\
                MRAEA & 51.32 & 56.89 & 72.67 & 77.97 & 74.65 & 79.96\\
                \midrule
                \bottomrule[2pt]
            \end{tabular}%
		}
	\end{center}
	\vspace*{-5mm}
\end{table}


\section{Conclusions and Future Directions}\label{sec:conclusion}

We have provided a comprehensive tutorial-type survey on representative EA techniques that use the new approach of representation learning. We have presented a framework for capturing the key characteristics of these techniques, proposed a benchmark DWY-NB to address the limitation of existing benchmark datasets, and conducted extensive experiments using the proposed datasets. The experimental study shows the comparative performance of the techniques and how various factors affect the performance. An insight from the experiments is that making good use of semantic information such as attribute triples improves the accuracy significantly. AttrE and MultiKE consistently perform the best in various settings of our experiments.

\textbf{Future Directions}. In terms of the benchmark, more experimental settings may be further explored such as varying the proportion of entities with the same name (i.e., the proportion of the ``tricky" feature), and the ratio between relation triples and attribute triples.

In terms of the accuracy of EA techniques, we may improve via pre-training. Pre-training has been very successful in NLP but its use in knowledge bases is limited to using pre-trained word embeddings for initializing entity name features. There is still huge potential of innovative ways of pre-training. For example, pre-trained predicate embeddings may be computed based on the predicate description to capture the semantics and similarity of predicates from different KGs. To train such embeddings, we may use transformer for relation prediction from relation descriptions, i.e., given a relation description, the model is trained to predict the corresponding relation. It may be further expanded into relation prediction between two entities, where the model takes the description of two entities and predicts the relations between the two entities. 


Various components and strategies used by EA techniques may be improved following the analysis and discussions based on our framework. First, many translation-based EA techniques uses TransE as the KG structure embedding (cf. Table~\ref{tab:model_summary}). We may explore replacing this component by improved versions of TransE such as TransD~\citep{TransD2015} and TransR~\citep{TransR2015}. Second, many translation-based techniques use the same loss function as TransE (Equation~\ref{eq:TransELoss}). We may try the limit-based loss function (Equation~\ref{eq:limit-loss}) which has been reported to have better performance~\citep{limit-loss2017}. 
%
%
Third, most of the existing EA techniques use seed alignments in the alignment module (cf. Figure~\ref{fig:framework}), but seed alignments are expensive and difficult to obtain, so unsupervised EA techniques will be an attractive direction. 

In terms of the efficiency and scalability of EA techniques, existing studies have mostly been conducted on small datasets. It is important to develop techniques that can conduct EA on very large KGs, improving on both efficiency and memory required.
\vspace{-0mm}



\begin{acknowledgements}
We would like to thank the anonymous reviewers whose comments have greatly helped improve this paper.
\end{acknowledgements}

%
%

\bibliographystyle{spbasic}      

\bibliography{references} 



\end{document}